\documentclass{article}
\usepackage{ijcai25}

\usepackage{times}
\usepackage{soul}
\usepackage{url}
\usepackage[hidelinks]{hyperref}
\usepackage[utf8]{inputenc}
\usepackage[small]{caption}
\usepackage{graphicx}
\usepackage{amsmath}
\usepackage{amsthm}
\usepackage{booktabs}
\usepackage{algorithm}
\usepackage{algorithmic}
\usepackage[switch]{lineno}
\usepackage{xcolor}
\usepackage{amssymb}
\usepackage{graphicx}
\usepackage{booktabs}
\usepackage{colortbl}
\usepackage{multirow}
\definecolor{mygray}{gray}{.9}

\newcommand{\best}[1]{{{\textbf{#1}}}}
\newcommand{\second}[1]{{\underline{{#1}}}}

\usepackage{xspace}
\newcommand{\NAMESSM}{Texture-Aware State Space Model\xspace}
\newcommand{\NAMENetWork}{TAMambaIR\xspace}

\usepackage[capitalize]{cleveref}
\crefname{section}{Sec.}{Secs.}
\Crefname{section}{Section}{Sections}
\Crefname{table}{Table}{Tables}
\crefname{table}{Tab.}{Tabs.}

\title{Directing Mamba to Complex Textures: An Efficient Texture-Aware State Space Model for Image Restoration}

\author{
  Long Peng$^{1,2\dag}$ \and Xin Di$^{1\dag}$\and Zhanfeng Feng$^{1}$ \and Wenbo Li$^{2}$ \and Renjing Pei${^{2*}}$  \and Yang Wang$^{1}$\thanks{$\dag$ is equal contribution. Renjing Pei, Yang Wang are the corresponding authors.}\and Xueyang Fu$^{1}$ \and Yang Cao$^{1}$ \and Zheng-Jun Zha$^{1}$ \\
\affiliations
  $^1$~University of Science and Technology of China \\
  $^2$~Huawei Noah’s Ark Lab \\
  \emails
  longp2001@mail.ustc.edu.cn, ywang120@ustc.edu.cn
}
\begin{document}

\maketitle

\begin{abstract}
Image restoration aims to recover details and enhance contrast in degraded images. With the growing demand for high-quality imaging (\textit{e.g.}, 4K and 8K), achieving a balance between restoration quality and computational efficiency has become increasingly critical. Existing methods, primarily based on CNNs, Transformers, or their hybrid approaches, apply uniform deep representation extraction across the image. However, these methods often struggle to effectively model long-range dependencies and largely overlook the spatial characteristics of image degradation (regions with richer textures tend to suffer more severe damage), making it hard to achieve the best trade-off between restoration quality and efficiency. To address these issues, we propose a novel texture-aware image restoration method, TAMambaIR, which simultaneously perceives image textures and achieves a trade-off between performance and efficiency. Specifically, we introduce a novel Texture-Aware State Space Model, which enhances texture awareness and improves efficiency by modulating the transition matrix of the state-space equation and focusing on regions with complex textures. Additionally, we design a {Multi-Directional Perception Block} to improve multi-directional receptive fields while maintaining low computational overhead. Extensive experiments on benchmarks for image super-resolution, deraining, and low-light image enhancement demonstrate that TAMambaIR achieves state-of-the-art performance with significantly improved efficiency, establishing it as a robust and efficient framework for image restoration.
\end{abstract}

\section{Introduction}
Image restoration as a fundamental task in computer vision and image processing, aiming to  recover details and improve image contrast from degraded images~\cite{a:22,a:37,MambaIR,zamir2022restormer,xiao2022image,yi2021efficient,yi2021structure,peng2020cumulative,wang2023decoupling,peng2021ensemble,wang2023brightness,he2024latent,peng2024unveiling}, which has been widely applied in imaging devices and various vision systems~\cite{SICE-Mix,2022LLE}. With the continuous advancements and widespread deployment of modern smartphones and cameras, the demand for higher image quality has significantly increased from the 1K resolutions (1280 $\times$ 720) to 4K (4096 $\times$ 2160)~\cite{Kong_2021_CVPR}. Therefore, balancing the growing demand for high-quality restoration quality and model efficiency has become a critical challenge in image restoration~\cite{liang2021swinir,MambaIR}.

\begin{figure}[t]
\centering
\includegraphics[width=0.5\textwidth]{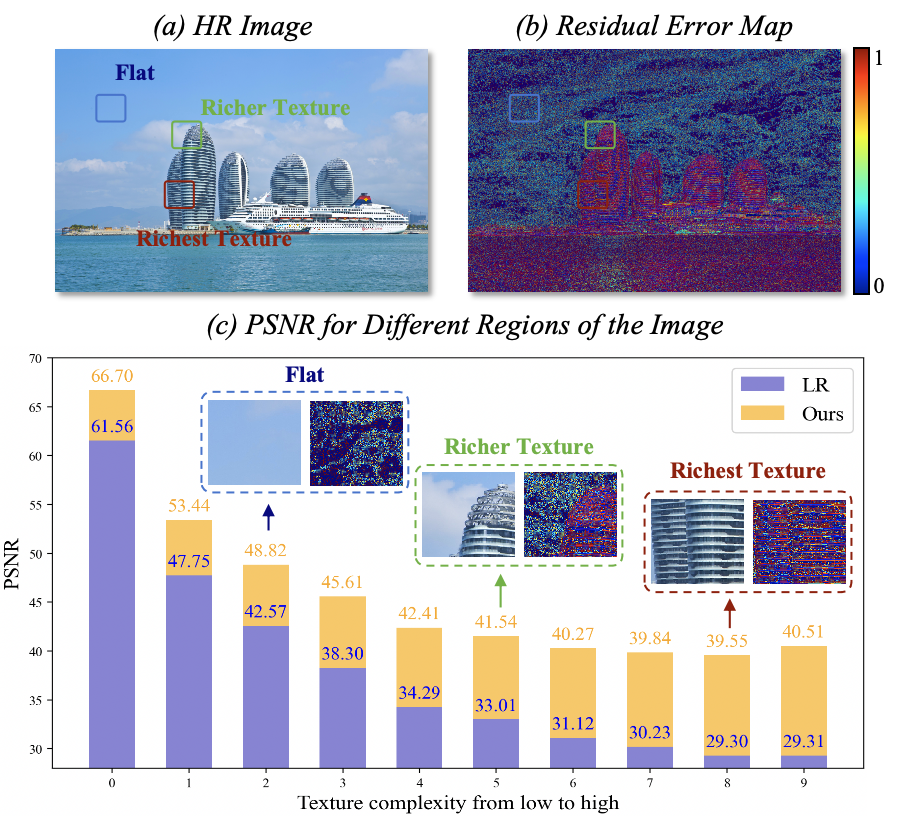}
\caption{(a-b) We compute the residual error map between the high-resolution and low-resolution images in the DIV2K and Manage109 datasets and find that the degree of degradation varies across different regions of the image. (c) We divide image patches into 10 groups sorted by texture complexity (measured by statistical variance) and calculate the average PSNR for each group in all Manage109 datasets. It can be observed that regions with richer textures suffer more severe degradation, resulting in lower PSNR values.
}
\label{fig:motivation}
\end{figure}

With the rapid advancement of deep learning, various image restoration methods have been proposed. Early research in image restoration primarily focused on convolutional neural networks (CNNs)~\cite{zhang2018rcan,sun2023spatially}, which offer high computational efficiency. However, limited by the local receptive fields of CNNs, these methods struggle to model long-range dependencies, resulting in suboptimal restoration quality. Therefore, Vision Transformers are introduced to significantly improve image restoration performance due to their ability to model long-range dependencies~\cite{Chen_2023_CVPR,zhou2023srformer,RetinexFormer}. Despite these advancements, the high computational complexity of Transformers remains a major obstacle for real-world deployment, especially for high-resolution images.

To seek efficient frameworks, many previous works have proposed combining the efficiency of CNN with the global modeling capabilities of Transformers, resulting in hybrid CNN-Transformer models that aim to balance performance and complexity~\cite{liang2021swinir,zhou2023srformer,fang2022hybrid}. However, these approaches often overlook the spatial characteristics of degradation, where regions with richer textures tend to suffer more severe damage, applying uniform deep representation extraction across the image. This uniform approach makes it challenging to achieve a trade-off between performance and efficiency. Specifically, we argue that regions with richer textures are generally more prone to severe quality degradation. To clarify this observation, we conduct a statistical analysis on image super-resolution (additional analyses of low-light image enhancement are shown in the appendix). First, we divide the low-resolution or low-quality images into patches and measure the texture complexity of each patch using statistical variance. We then evaluate the PSNR between the degraded patches and their ground truth counterparts. The results, as shown in Figure~\ref{fig:motivation}, reveal that regions with richer textures tend to suffer more severe degradation, resulting in lower PSNR values. This highlights the importance of paying more attention to regions with complex textures. Although these methods~\cite {Kong_2021_CVPR,jeong2025accelerating} attempt to address this issue by classifying image patches/pixels and applying different-sized CNNs, they still face challenges in perceiving image textures and modeling contextual information, making it difficult to achieve high-efficiency.

To address these issues, we propose a novel \NAMESSM (TA-SSM) to simultaneously achieve efficient contextual modeling and texture awareness, enabling an optimal trade-off between performance and efficiency. Specifically, we modulate the transition matrix of state-space equations to mitigate catastrophic forgetting in regions with richer textures, enhancing texture perception. Furthermore, by focusing more on challenging texture regions, the proposed method improves overall efficiency. Additionally, for the first time, we introduce positional embeddings into SSM, improving their ability to perceive contextual positions. To further reduce computational costs while maintaining multi-directional perception, we design a Multi-Directional Perception Block to enhance efficiency. Extensive experiments on various image restoration benchmarks including image super-resolution, image deraining, and low-light image enhancement, demonstrate that the proposed method significantly improves the efficiency of Mamba-based image restoration. This provides a more powerful and efficient framework for future image restoration tasks.

{The contributions of this paper can be summarized:}
\begin{itemize}
\item {We propose a novel and efficient \NAMESSM (TA-SSM) that perceives complex textures by modifying the state-space equations and transition matrices and simultaneously focuses on more challenging texture regions to enhance efficiency.}

\item {An efficient Multi-Directional Perception Block is proposed to expand the receptive field in multiple directions while maintaining low computational costs. Furthermore, position embedding is introduced into SSM  to enhance its capability of perceiving contextual positions.}

\item {Based on these, a straightforward yet efficient model \NAMENetWork is proposed, showcasing superior performance in both efficiency and effectiveness across various image restoration tasks and benchmarks, offering an efficient backbone for image restoration.}
\end{itemize}

\section{Related work}

\subsection{Image Restoration}
{Images captured in complex real-world scenarios often suffer from degradations like low resolution, low-light, rain, and haze, resulting in reduced contrast and detail loss~\cite{RealBlur,wang2018esrgan,a:35,a:36,peng2024lightweight,peng2024towards,peng2024efficient}. Image restoration aims to enhance contrast and recover details, improving visual quality. Advances in deep learning have significantly boosted its effectiveness~\cite{liang2021swinir}. Dong \textit{et al.}~\cite{dong2015image} introduced SRCNN in 2015, pioneering deep learning for image super-resolution (SR). Since then, numerous CNN-based methods have emerged to address image restoration tasks~\cite{zhang2018rcan,dai2019second}, though their lightweight designs often limit receptive fields. To address this issue, Vision Transformers (ViTs) have been introduced to leverage long-range modeling for image restoration. Chen \textit{et al.}~\cite{chen2021pre} demonstrated their effectiveness in image denoising, while the Swin Transformer~\cite{liang2021swinir} captured multi-scale features hierarchically. However, increasing image resolutions pose significant computational challenges for Transformer-based methods~\cite{MambaIR}. While CNN-Transformer hybrids aim to balance performance and efficiency, they often overlook the spatial characteristics of degradation, where regions with richer textures tend to suffer more severe damage. As a result, they apply uniform enhancements and struggle to effectively focus on the damaged textures. Patch- or pixel-level classification methods~\cite{Kong_2021_CVPR,jeong2025accelerating} address this partially but fail to effectively model textures and contextual information. To tackle this, we adopt efficient global modeling with State Space Model (SSM) and introduce texture-aware capabilities to better handle complex textures, achieving a balance between performance and efficiency.
\begin{figure*}[t]
\centering
\includegraphics[width=1.0\textwidth]{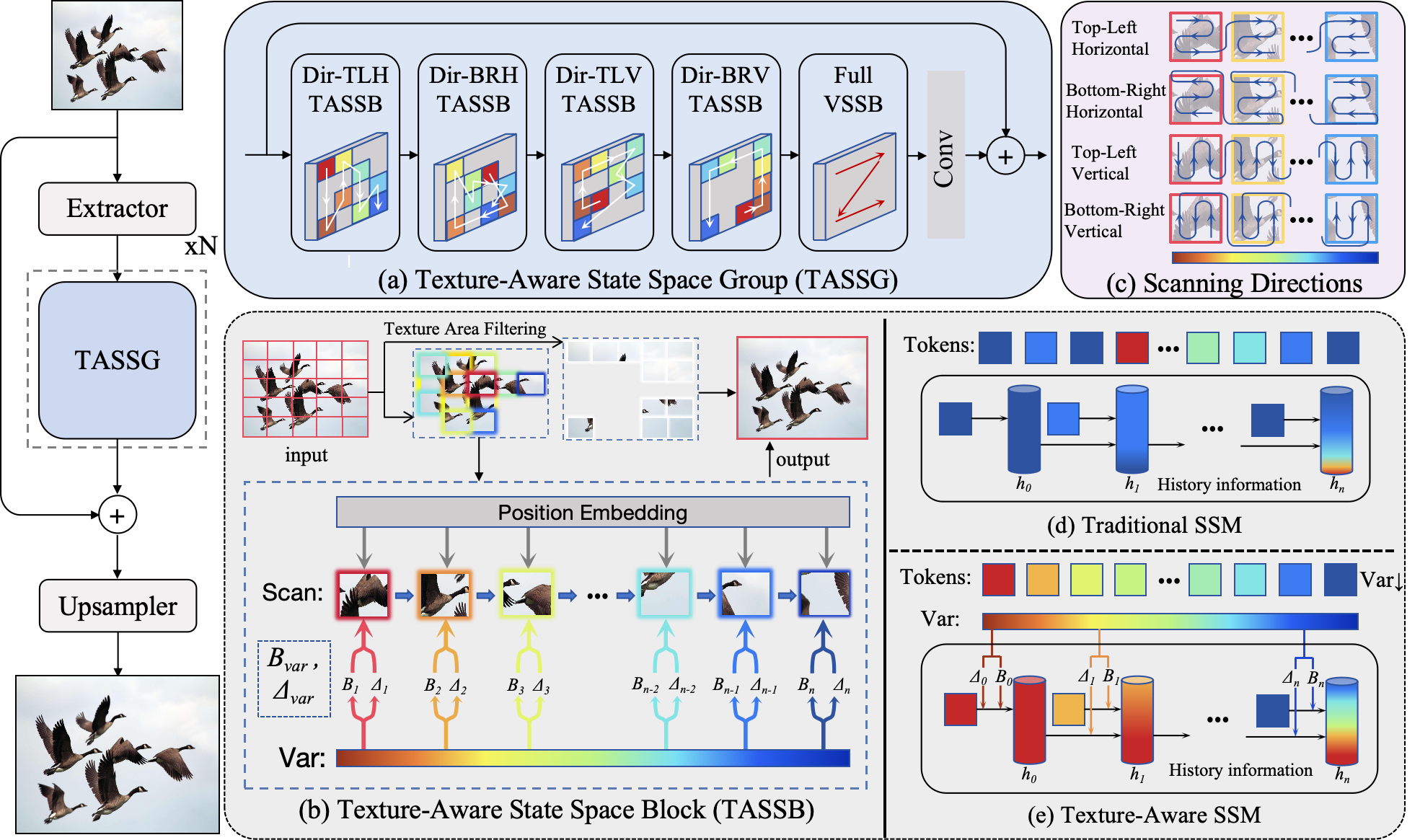}
\caption{The overall framework of \NAMENetWork consists of a feature extractor and several TASSGs, with an upsampler by pixel shuffle.}
\label{fig:framework}
\end{figure*}
\subsection{State Space Model}
State Space Model (SSM), originally developed in the 1960s for control systems~\cite{a:25}, has recently been extended to applications in computer vision~\cite{a:30,patro2024simba,fu2024ssumamba,chen2024changemamba}. The introduction of SSM into computer vision was pioneered by Visual Mamba, which designed the VSS module to achieve superior performance compared to Vision Transformers (ViTs)~\cite{a:31}, while maintaining lower model complexity. This breakthrough inspired a wave of research utilizing SSM in various tasks~\cite{c:8,a:32,a:33,a:34,a:35,a:36}. Notably, MambaIR~\cite{c:8} is the first to employ SSM for image restoration, demonstrating improved efficiency and enhanced global perceptual capabilities. However, directly applying SSM to image restoration poses challenges in effectively enhancing texture-rich regions, which limits the full potential of the State Space Model in this domain.

\section{Proposed Method}

\subsection{Traditional State Space Model}
Let's briefly review the traditional State Space Model (SSM), which maps sequence input \( x(t) \) to output \( y(t) \) through an implicit latent state \( h(t) \in \mathbb{R}^N \)~\cite{c:8} and can be represented as a linear ordinary differential equation:
\begin{equation}
\begin{aligned}
    \dot{h}(t) &= Ah(t) + Bx(t), \\
    y(t) &= Ch(t) + Dx(t),
\end{aligned}
\label{eq:1}
\end{equation}
where \(N\) is the state size, \(A \in \mathbb{R}^{N \times N}\), \(B \in \mathbb{R}^{N \times 1}\), \(C \in \mathbb{R}^{1 \times N}\), and \(D \in \mathbb{R}\). Discretized using a zero-order hold as:
\begin{equation}
\begin{aligned}
    \overline{A} &= \exp(\Delta A), \\
    \overline{B} &= (\Delta A)^{-1} (\exp(\Delta A) - I) \Delta B,
\end{aligned}
\label{eq:2}
\end{equation}
After the discretization, the discretized version of Eq.~\eqref{eq:1} with step size \(\Delta\) can be rewritten as:
\begin{equation}
\begin{aligned}
    h_k &= \overline{A} h_{k-1} + \overline{B} x_k, \\
    y_k &= Ch_k + Dx_k,
\end{aligned}
\label{eq:3}
\end{equation}
The advent of SSM has garnered significant attention for its ability to perform global modeling with linear complexity in image restoration~\cite{MambaIR}. However, traditional SSM still faces two critical challenges: \textbf{(1) It consistently handles flat and texture-rich regions, spending too much computation on easily recoverable flat areas (as shown in Figure~\ref{fig:motivation}), resulting in inefficiency. (2) It lacks an effective texture-awareness mechanism, which not only leads to catastrophic forgetting of texture information but also hinders the model's ability to focus on challenging texture-rich regions.} 


\subsection{\NAMESSM}
To address these limitations, we propose a novel \NAMESSM to fully unlock the potential of SSM in image restoration, as shown in Figure~\ref{fig:framework}. We tackle the aforementioned challenges from two perspectives. First, we propose a novel Texture Area Filtering method, which utilizes variance statistics to identify texture-rich and flat regions and pay less/more computation on flat/texture-rich regions, thereby improving efficiency. Second, we propose a novel Texture-Aware Modulation, which ranks regions based on their texture richness in descending order and enhances SSM's historical retention of texture-rich information by modulating the transition matrix in the state-space equation.

\noindent  \textbf{Texture Area Filtering.} We first patchify the image/feature \( x \in \mathbb{R}^{ C \times H \times W} \) into patches \( p \in \mathbb{R}^{ C \times h \times w} \) to obtain a sequence \( \{P_i\} \), as follows: 
\begin{equation}
\{P_i\} = {Patchify}(x), \quad i = 1, 2, \ldots, N .
\end{equation}
where $N$ represents the number of patches. Then, we sort this patch sequence based on the texture complexity from high to low, which is measured by statistical variance (other measurements of compression are provided in the appendix), resulting in a new ordered patch sequence \( \{P^{'}_i\} \), as follows:
\begin{equation}
\begin{aligned}
\mu(P_i) &= \frac{1}{|P_i|} \sum_{j \in P_i} P_{ij}, \\ 
\mathrm{Var}(P_i) &= \frac{1}{|P_i|} \sum_{j \in P_i} \left(P_{ij} - \mu(P_i)\right)^2, \\
\{P_i'\}_{i=1}^n &= \operatorname{argsort}_\downarrow \big(\mathrm{Var}(P_i)\big),
\end{aligned}
\end{equation}
Then, we process the top \( p\% \) patches with the highest texture complexity and skip the easy flat regions (which will be enhanced by the following full-sequence scan and convolution as shown in the (a) of Figure~\ref{fig:framework} ),  directing SSM focus on the more challenging region as follows:
\begin{equation}
\{P_i\}_{{top-}p\%} = \big\{P_i' \mid i \leq \lceil p \cdot N \rceil \big\},
\label{eq:6}
\end{equation}

\noindent  \textbf{Texture-Aware Modulation} In traditional SSM, the transition matrixes $B$ and $\Delta$ are expressed as follows:
\begin{equation}
B=Linear(x),\Delta = Linear(x).
\end{equation}
This way may lead to two key issues: (a) The transition matrix relies solely on the input itself and lacks the ability to perceive texture information, \textbf{resulting in catastrophic forgetting of texture-related representations in the state-space history, thereby weakening the model's ability to enhance fine details}, as shown in the  (d) of Figure~\ref{fig:framework}. (b) Flat regions, which are easier to enhance, dominate the state-space history, \textbf{leading to an excessive focus on flat areas and hindering the model's ability to address challenging texture-rich regions}. To address these issues and enhance texture awareness, we first sort the patches as described in Eq.\ref{eq:6}, and then propose modulating the transition matrix within the state-space model to improve its capability of preserving historical texture information, as shown in Figure\ref{fig:framework} (b) and (e).
\begin{equation}
\begin{aligned}
B_\text{var}&=Var(x)\cdot Linear(x), \\
\Delta_\text{var} &= Var(x)\cdot Linear(x).
\end{aligned}
\end{equation}
Through this method, the transition matrices $B_{var}$ and $\Delta_{car}$ are able to capture the varying texture complexities of different patches. Furthermore, they retain the historical information of regions with richer textures, effectively alleviating catastrophic forgetting and enhancing the ability to represent fine details in image restoration, as illustrated in Figure~\ref{fig:framework} (e). Finally, the state-space function formulation of \NAMESSM can be expressed as follows:
\begin{equation}
\begin{aligned}
\overline{A}_{\text{var}_k} &= \exp(\Delta_{\text{var}_k} A) \\
\overline{B}_{\text{var}_k} &= (\Delta_{\text{var}_k} A)^{-1} (\exp(A) - I) \cdot \Delta_{\text{var}_k} B_{\text{var}_k} \\
h_k &= \overline{A}_{\text{var}_k} h_{k-1} + \overline{B}_{\text{var}_k} x_k \\
y_k &= C h_k + D x_k.
\end{aligned}
\end{equation}

\noindent \textbf{Position Embedding.} It is well known that the traditional SSM adopts a linear sequence scanning method, which inevitably limits the contextual receptive field. Although SSM proposes scanning in four directions to partially improve the contextual receptive field, the perception capabilities of contextual positions remain limited. On the other hand, considering that our TA-SSM employs a texture complexity-based scanning method, this further exacerbates the contextual position perception process. To address this, we propose to introduce position embedding, which is widely used in Transformers, into SSM to enhance its contextual position awareness. This method enables SSM to capture the global positional relationship of the current token during each sequence scan, thereby improving its ability to perform global texture arrangement and contextual modeling. Specifically, we add a learnable position embedding $Pos$ to each patch after Eq.~\ref{eq:6}:
\begin{equation}
\{P_i\}_{{top-}p\%} = \{P_i + Pos(P_i) \mid P_i \in \{P_i\}_{{top-}p\%}\}.
\end{equation}

\subsection{Multi-Directional Perception Block}
\label{sec:MDPB}
Through \NAMESSM can model the global information of different patches based on their texture complexity. However, the contextual perception within a patch remains critical. Existing approaches often adopt multi-directional scanning within a single SSM, which can lead to significant computational overhead. To address this issue, we propose a novel Multi-Directional Perception Block that sequentially connects four different scaning directions (Top-Left 
Horizontal, Bottom-Right Horizontal, Top-Left Vertical, and Bottom-Right Vertical) to reduce computational cost while enhancing multi-directional perception, as shown in Figure~\ref{fig:framework} (a) and (c). Additionally, low-texture complexity regions (\textit{e.g.}, flat areas), which are relatively easier to enhance, are also crucial for image restoration. Therefore, we introduce a full-scan SSM and convolution layer to ensure the network can also focus on these regions, facilitating interactions between high-texture complexity patches and flat patches.



\begin{table*}[ht]
\centering
\vspace{-2mm}
\caption{Quantitative comparison on \textbf{$\times 2$ image super-resolution} with state-of-the-art methods. The best and the second best results are in \best{bold} and \second{bold}.}
\label{tab:classicSR}
\scalebox{0.95}{

\begin{tabular}{@{}l|cc|cc|cc|cc|c|c@{}}
\toprule
 & \multicolumn{2}{c|}{\textbf{Set5}} &
  \multicolumn{2}{c|}{\textbf{Set14}} &
  \multicolumn{2}{c|}{\textbf{BSDS100}} &
  \multicolumn{2}{c|}{\textbf{Manga109}} & \textbf{FLOPs (G)} & \textbf{Params (M)} \\
\multirow{-2}{*}{Method} & PSNR$\uparrow$  & SSIM$\uparrow$   & PSNR$\uparrow$  & SSIM$\uparrow$   & PSNR$\uparrow$  & SSIM$\uparrow$   & PSNR$\uparrow$  & SSIM$\uparrow$   &  &  \\ \midrule
RCAN   & 38.27 & 0.9614 & 34.12 & 0.9216 & 32.41 & 0.9027 & 39.44 & 0.9786 & 62.75 & 15.44 \\
SAN    & 38.31 & 0.9620 & 34.07 & 0.9213 & 32.42 & 0.9028 & 39.32 & 0.9792 & 64.11 & 15.71 \\
ClassSR   & 38.29 & 0.9615 & 34.18 & 0.9218 & 32.45 & 0.9029 & 39.45 & 0.9786 & 40.78 & 30.10 \\
IPT    & 38.37 & - &34.43 &- & 32.48&- & -& - & - & 115.61 \\
CSNLN  & 38.28 & 0.9616 & 34.12 & 0.9223 & 32.40 & 0.9024 & 39.37 & 0.9785 & 481.97 & 6.21 \\
NLSA   & 38.34 & 0.9618 & 34.08 & 0.9231 & 32.43 & 0.9027 & 39.59 & 0.9789 & 182.82 & 41.80 \\
EDT-B   & 38.45 & 0.9624 & \second{34.57} & \second{0.9258} & 32.52 & 0.9041 & 39.93 & 0.9800 & 30.22 & 11.48 \\
EDSR   & 38.11 & 0.9602 & 33.92 & 0.9195 & 32.32 & 0.9013 & 39.10 & 0.9773 & 166.84 & 40.73 \\
RDN   & 38.24 & 0.9614 & 34.01 & 0.9212 & 32.34 & 0.9017 & 39.18 & 0.9780 & 90.60 & 22.12 \\
HAN    & 38.27 & 0.9614 & 34.16 & 0.9217 & 32.41 & 0.9027 & 39.46 & 0.9785 & 258.82 & 63.61 \\
SwinIR & 38.42 & 0.9623 & {34.46} & 0.9250 & 32.53 & 0.9041 & 39.92 & 0.9797 & 51.33 & 11.75 \\
SRFormer &  \second{38.51} &\second{0.9627}&34.44&{0.9253}&\second{32.57} &\second{0.9046} &\second{40.07} &0.9802 & 62.95 & 10.40 \\

\NAMENetWork-S &  \best{38.53} &\best{0.9627}&\best{34.64}&\best{0.9262}&\best{32.57} &\best{0.9046} &\best{40.23} &\best{0.9806} & 56.88 & 12.19 \\

\midrule 
MambaIR & \second{38.57}&\second{0.9627}&\second{34.67}&\second{0.9261}&\second{32.58}&\second{0.9048}&\second{40.28}&\second{0.9806} & 110.49 & 20.42 \\ 
\NAMENetWork & \best{38.58} & \best{0.9627} & \best{34.72} & \best{0.9265} & \best{32.58} & \best{0.9048} & \best{40.35} & \best{0.9810} & 89.99 & 16.07\\ 
\bottomrule
\end{tabular}%
}
\end{table*}

\subsection{\NAMENetWork}
Following the previous approach in~\cite{MambaIR}, we adopt a simple yet efficient architecture to build \NAMENetWork, as depicted in Figure~\ref{fig:framework}. \NAMENetWork comprises a feature extractor, a Texture-Aware State Space Group, and an upsampler. Specifically, convolution layers and pixel shuffle are used in the feature extractor and upsampler. Due to limited space, the implement details of the Texture-Aware State Space Group and Texture-Aware State Space Block are provided in \textbf{Appendix Section 8}.

\subsection{Loss Function}

Following previous works~\cite{zamir2022restormer,focal}, we utilize the L1 loss \(\mathcal{L}_{1}\) and frequency loss \(\mathcal{L}_{fft}\)} for training. The total loss is presented as follows:
\begin{equation}
\label{loss}
\mathcal{L}_{total} = \lambda_{1} \mathcal{L}_{1}(\mathcal{O}, \mathcal{O}_{gt}) + \lambda_{2} \mathcal{L}_{fft}(\mathcal{O}, \mathcal{O}_{gt}) .
\end{equation}
where $\mathcal{O}$ and $\mathcal{O}_{gt}$ denote the model output and the ground truth, respectively. The parameters $\lambda_{1}$ and $\lambda_{2}$ are balancing factors. We set $\lambda_{1}$ and $\lambda_{2}$ to 1, 0.05, respectively.

\section{{Experiences}}

\subsection{Experimental Settings}

\noindent \textbf{{Training Details.}} Following prior works~\cite{liang2021swinir,MambaIR}, the training batch sizes for image super-resolution, image deraining, and low-light image enhancement are set to 32, 16, and 16, respectively. During training, the original images are cropped into $64 \times 64$ patches for image super-resolution, and $256 \times 256$ patches for image deraining and low-light image enhancement. Top \( p\% \) epirically is set at 0.5. Following~\cite{liang2021swinir,MambaIR}, we use the UNet architecture for image deraining and low-light image enhancement. The Adam optimizer~\cite{kingma2014adam} is used to train our method, with $\beta_1 = 0.9$, $\beta_2 = 0.999$, and an initial learning rate of $2 \times 10^{-4}$. The standard model consists of 7 intermediate blocks with a depth of 7, while the small version consists of 6 intermediate blocks with a depth of 6. All experiments are conducted on 8 NVIDIA V100 GPUs. Additional details of the implementation are provided in \textbf{Appendix Section 9}.

\noindent \textbf{Dataset.} To evaluate our method, we select three popular and representative image restoration tasks, including image super-resolution (SR), image draining (Derain), and low-light image enhancement (LLIE). Specifically, for SR, DIV2K~\cite{timofte2017ntire} and Flickr2K~\cite{lim2017enhanced} are employed to train the network, while Set5~\cite{bevilacqua2012low}, Set14~\cite{zeyde2012single}, B100~\cite{martin2001database}, and Manga109~\cite{matsui2017sketch} are used for evaluation. For Derain, we follow~\cite{DRSformer} and validate our approach on both popular benchmarks: Rain200H and Rain200L~\cite{yang2017deep}, which contain heavy and light rain conditions, respectively, for training and evaluation. For LLIE, we follow~\cite{cai2023retinexformer} and validate our approach on the synthetic version of the LOL-V2 dataset~\cite{LOLv2}. More details about the datasets used are provided in the \textbf{Appendix Section 10}.

\noindent \textbf{Evaluation Metrics.} Following previous work~\cite{DRSformer,cai2023retinexformer,MambaIR}, we use Peak Signal-to-Noise Ratio (PSNR) and Structural Similarity Index (SSIM) as evaluation metrics. For the image super-resolution and deraining tasks, evaluation is performed on the Y channel of the YCbCr color space, while for the low-light enhancement task, evaluation is conducted in the RGB space.

\noindent{\textbf{Comparisons with State-of-the-art Methods.} We compare with many existing state-of-the-art methods. \textbf{{Image super-resolution}}: we compare our method against thirteen state-of-the-art methods, including: RCAN, SAN, ClassSR, IPT, CSNLN, NLSA, EDT-B, EDSR, RDN, HAN, SwinIR, SRFormer, and MambaIR. \textbf{{Image deraining}}: we compare our method against fifteen state-of-the-art methods, including: DDN, RESCAN, PReNet, MSPFN, RCDNet, MPRNet, SwinIR, DualGCN, SPDNet, Uformer, Restormer, IDT, DLINet, DRSformer, and MambaIR. \textbf{{Low-light image enhancement}}: we compare our method against fifteen state-of-the-art methods, including: RetinexNet, KinD, ZeroDCE, 3DLUT, DRBN, RUAS, LLFlow, EnlightenGAN, Restormer, LEDNet, SNR-Aware, LLFormer, RetinexFormer, CIDNet, and MambaIR. Due to the limited space, more detail and references are provided in \textbf{Appendix Section 11}.
}


\begin{table}[t]
\caption{Quantitative comparison on {\textbf{low-light image enhancement}} with state-of-the-art methods.}
\centering
\label{tab:lowlightenhancement}
\scalebox{0.9}{

\begin{tabular}{l|cccc}
\toprule
\multicolumn{1}{c|}{\cellcolor[HTML]{FFFFFF}}                          & \multicolumn{2}{c}{\textbf{Normal}}                                                     & \multicolumn{2}{c}{\textbf{GT Mean}}                                                    \\
\multicolumn{1}{c|}{\multirow{-2}{*}{Methods}} & PSNR$\uparrow$                                   & SSIM$\uparrow$                                  & PSNR$\uparrow$                                   & SSIM$\uparrow$                                  \\ \midrule
RetinexNet                                                    & 17.137                                 & 0.762                                 & 19.099                                 & 0.774                                 \\
KinD                                                         & 13.290                                 & 0.578                                 & 16.259                                 & 0.591                                 \\
ZeroDCE                                                      & 17.712                                 & 0.815                                 & 21.463                                 & 0.848                                 \\
3DLUT                                                        & 18.040                                 & 0.800                                 & 22.173                                 & 0.854                                 \\
DRBN                                                         & 23.220                                 & 0.927                                 & -                                      & -                                     \\
RUAS                                                         & 13.765                                 & 0.638                                 & 16.584                                 & 0.719                                 \\
LLFlow                                                       & 24.807                                 & 0.919                                 & 27.961                                 & 0.930                                 \\
EnlightenGAN                                                 & 16.570                                 & 0.734                                 & -                                      & -                                     \\
Restormer                                                    & 21.413                                 & 0.830                                 & 25.428                                 & 0.859                                 \\
LEDNet                                                       & 23.709                                 & 0.914                                 & 27.367                                 & 0.928                                 \\
SNR-Aware                                                    & 24.140                                 & 0.928                                 & 27.787                                 & 0.941                                 \\
LLFormer                                                     & 24.038                                 & 0.909                                 & 28.006                                 & 0.927                                 \\
RetinexFormer                                                & 25.670                                 & 0.930                                 & 28.992                                 & 0.939                                 \\
CIDNet                                                       & 25.705                                 & 0.942                                 & 29.566                                 & 0.950                                 \\
MambaIR                                                       & \second{25.830}                                 & \second{0.953}                                 & \second{30.445}                                 & \second{0.957}                               \\
\NAMENetWork                                                         & {\color[HTML]{000000} \textbf{26.735}} & {\color[HTML]{000000} \textbf{0.951}} & {\color[HTML]{000000} \textbf{31.358}} & {\color[HTML]{000000} \textbf{0.961}} \\ \bottomrule

\end{tabular}

}
\end{table}

\begin{table}[t]
\centering
\caption{Quantitative comparison on {\textbf{image deraining}} with state-of-the-art methods.}
\label{tab:Derain}
\scalebox{0.9}{

\begin{tabular}{l|cccc}
\toprule
                                 & \multicolumn{2}{c}{Rain200L} & \multicolumn{2}{c}{Rain200H} \\
\multicolumn{1}{c|}{\multirow{-2}{*}{Methods}}                        & PSNR$\uparrow$         & SSIM $\uparrow$          & PSNR $\uparrow$         & SSIM $\uparrow$        \\ \midrule
DDN                            & 34.68        & 0.9671        & 26.05        & 0.8056       \\
RESCAN                         & 36.09        & 0.9697        & 26.75        & 0.8353        \\
PReNet                        & 37.80         & 0.9814        & 29.04        & 0.8991        \\
MSPFN                          & 38.58        & 0.9827        & 29.36        & 0.9034       \\
RCDNet                          & 39.17        & 0.9885        & 30.24        & 0.9048        \\
MPRNet                         & 39.47        & 0.9825        & 30.67        & 0.9110        \\
SwinIR    & 40.61 & 0.9871 & 31.76 & 0.9151 \\
DualGCN  & 40.73        & 0.9886        & 31.15        & 0.9125        \\
SPDNet    & 40.50         & 0.9875        & 31.28        & 0.9207        \\
Uformer & 40.20         & 0.9860         & 30.80         & 0.9105        \\
Restormer & 40.99        & 0.9890         & 32.00           & 0.9329        \\
IDT     & 40.74        & 0.9884        & 32.10         & \underline{0.9344}        \\
DLINet                         & 40.91        & 0.9886        & 31.47        & 0.9231        \\
DRSformer & \underline{41.23}        & 0.9894        & 32.17        & 0.9326        \\
MambaIR & 41.13        & \underline{0.9895}        & \underline{32.18}        & 0.9295        \\
\NAMENetWork          & \textbf{41.25}           & \textbf{0.9896}                & \textbf{32.19}             & \textbf{0.9345}              \\
\bottomrule            
\end{tabular}

}
\end{table}

\begin{figure*}[ht]
\centering
\includegraphics[width=1.0\textwidth]{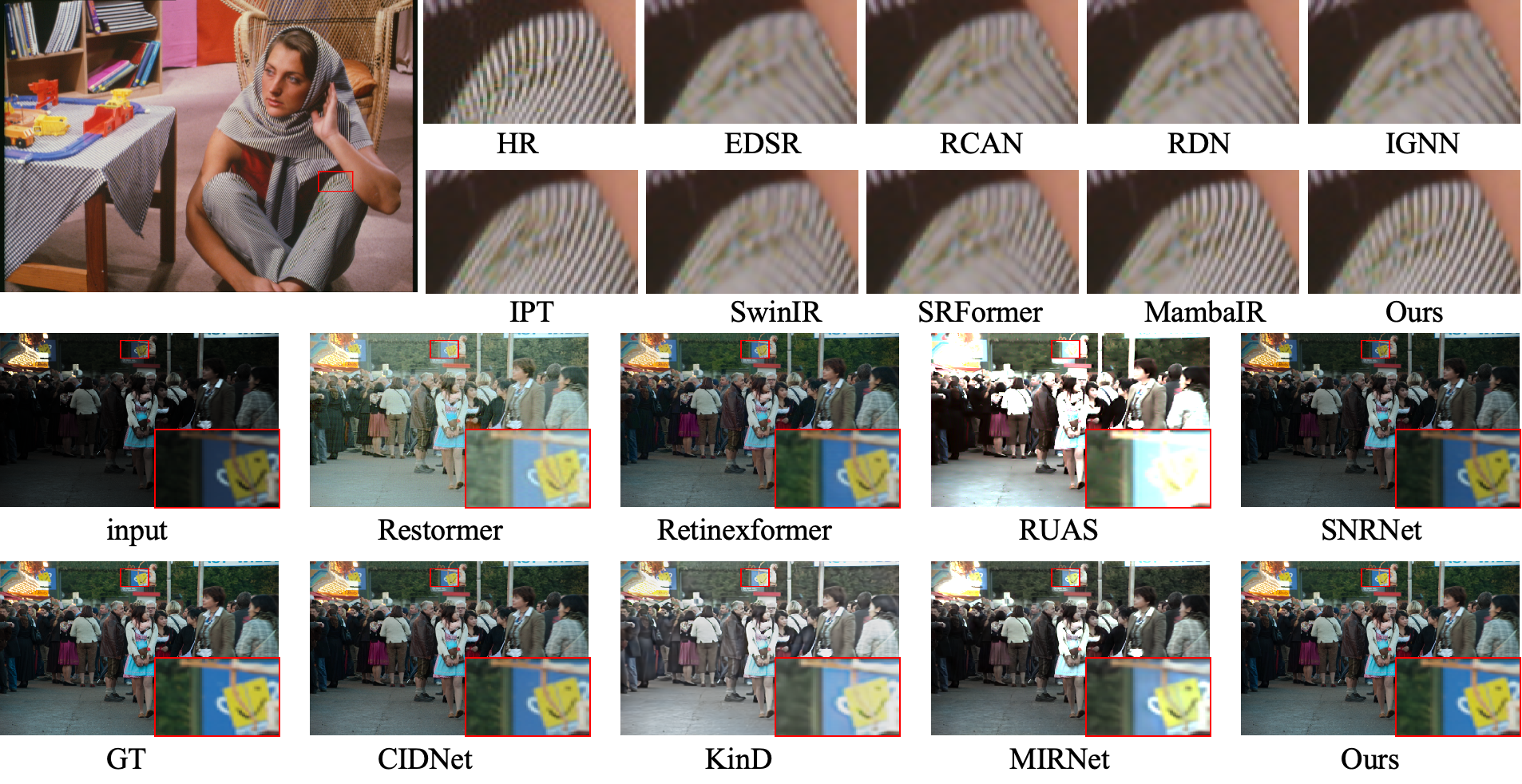}
\caption{Qualitative comparison on image super-resolution and low-light image enhancement.}
\label{fig:vis_comp}
\end{figure*}

\subsection{Ablation Study}
Table~\ref{tab:ablation} presents the results of our ablation study on \NAMENetWork-S at Manga109 datasets, showing the impact of removing critical components from the model. Excluding positional embeddings (w/o PosEmb) reduces the PSNR to 40.11, indicating their importance in enhancing the model's ability to perceive and utilize spatial context, which is especially beneficial for accurately restoring global structures. Removing the Multi-Directional Perception Block (w/o MDPB) and using a single direction to replace it reduces the PSNR to 40.14, demonstrating its role in capturing features from multiple directions efficiently. Replacing TA-SSM with traditional SSM and keeping the similar complexity causes the PSNR to decline to 39.98. This emphasizes the crucial role of TA-SSM in prioritizing complex textures and effectively modeling challenging regions. The complete model achieves the highest PSNR of 40.23, underscoring the necessity and complementary contributions of all components in delivering optimal performance. More ablation studies are provided in \textbf{Appendix Section 12}.

\subsection{{Quantitative Results}}
\noindent \textbf{Image Super-Resolution.}  We compare our method against thirteen state-of-the-art CNN-based, transformer-based, and Mamba-based SR approaches. As shown in Table~\ref{tab:classicSR}, our method achieves the best performance across all four benchmarks. Specifically, the small version of our method surpasses SRFormer by 0.15 dB and 0.20 dB on the PSNR of Manga109 and Set14 datasets, respectively. Moreover, the standard version of our method outperforms the current state-of-the-art method, MambaIR, with PSNR improvements of 0.07 dB and 0.05 dB on the Manga109 and Set14 datasets, respectively. These results highlight the superior performance and effectiveness of our approach.

\noindent \textbf{Model Complexity Comparison.} As shown in Table~\ref{tab:classicSR}, benefiting from its strong texture-awareness capability, our method not only achieves the best performance but also demonstrates significant advantages in computational efficiency, including FLOPs (with input size 64 $\times$ 64) and model parameters. Specifically, the small version of our model \NAMENetWork-S achieves state-of-the-art performance with relatively low complexity, surpassing SRFormer in terms of performance while reducing FLOPs by 6.07G. Furthermore, the standard version of our network outperforms MambaIR while achieving reductions of 20.5G (18.5\%) in FLOPs and 4.35M (21.3\%) in parameters. These results highlight the efficiency of our method. More comparisons of inferencing time are provided in \textbf{Appendix Section 13}.

\noindent \textbf{Image Deraining.} We compare our method against fifteen state-of-the-art deraining approaches, including CNN-based, transformer-based, and Mamba-based methods. As shown in Table~\ref{tab:Derain}, our method achieves the best performance on both Rain200L and Rain200H datasets in terms of PSNR and SSIM. Specifically, our method achieves a PSNR of 41.25 dB and an SSIM of 0.9896 on Rain200L, surpassing DRSformer, the previous best method, by 0.02 dB and 0.002, respectively. Similarly, on Rain200H, our approach attains a PSNR of 32.19 dB and an SSIM of 0.9345, outperforming MambaIR by 0.01 dB and 0.005, respectively.

\noindent \textbf{Low-Light Image Enhancement.} We compare our method with thirteen state-of-the-art approaches, including both CNN-, Transformer- and Mamba-based methods. We follow the Normal and GT Mean test settings in~\cite{CIDNet}, and the results are shown in Table~\ref{tab:lowlightenhancement}. Our method achieves the best performance on all settings in terms of PSNR and SSIM. Specifically, under the Normal setting, our method achieves a PSNR of 26.735 dB and an SSIM of 0.951, significantly surpassing RetinexFormer by 1.03 dB in PSNR and 0.021 in SSIM.  Under the GT Mean setting, our method achieves a PSNR of 31.358 dB and an SSIM of 0.961, outperforming MambaIR by a notable margin. These results demonstrate that our approach effectively enhances low-light images while preserving details.

\begin{table}[t]
\centering
\caption{Ablation studies on our proposed core module.}
\footnotesize
\label{tab:ablation}

\begin{tabular}{c|cccc}
\toprule
 & w/o PosEmb & w/o MDPB & w/o TA-SSM & Ours   \\ \midrule
PSNR                                                    & 40.11         & 40.14        & 39.98       & 40.23  \\
SSIM                                                    & 0.9801        & 0.9802       & 0.9796      & 0.9806 \\ \bottomrule
\end{tabular}


\end{table}

\subsection{{Qualitative Results}}
\begin{figure}[t]
\centering
\vspace{-2mm}
\includegraphics[width=0.45\textwidth]{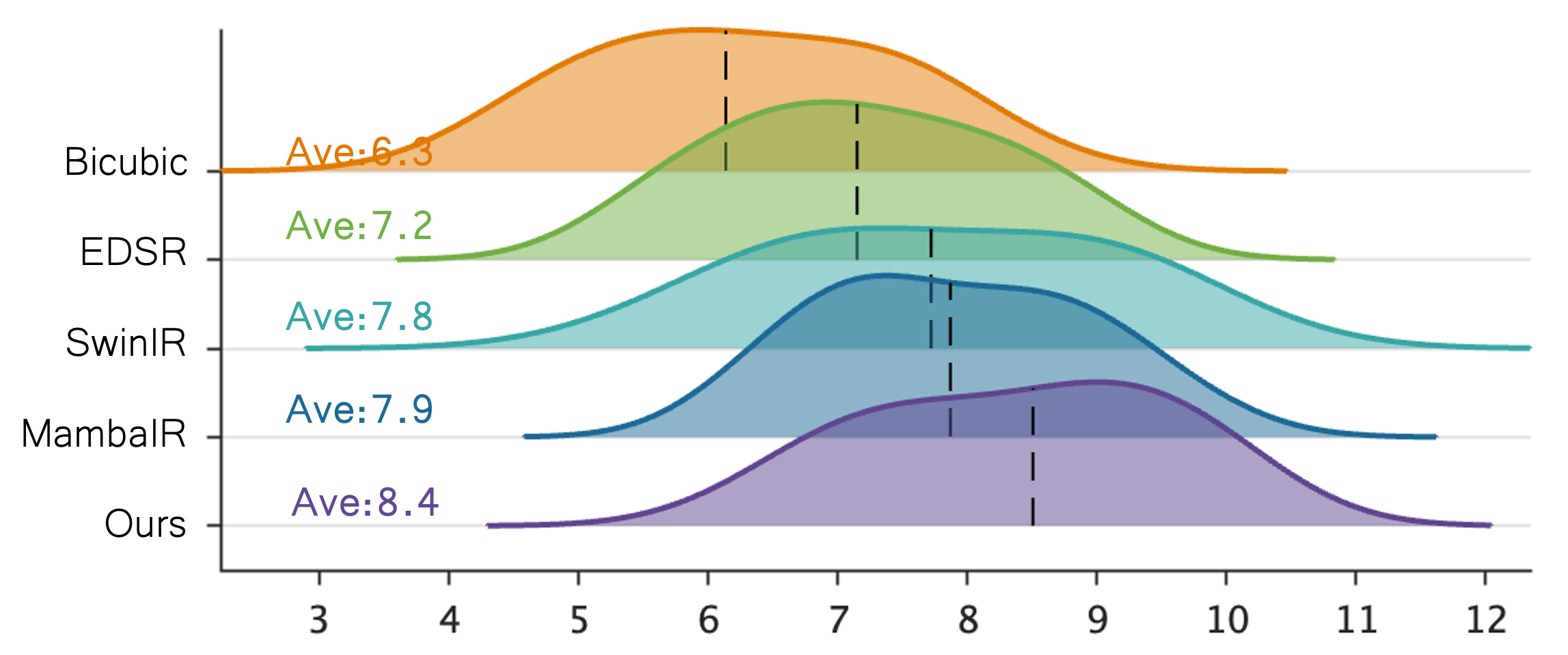}
\caption{User study on image visual quality.}
\vspace{-2mm}
\label{fig:user_study}
\end{figure}

We present qualitative comparisons to demonstrate the effectiveness of our method on both image super-resolution (SR) and low-light image enhancement (LLIE), as shown at the top of  Figure~\ref{fig:vis_comp}. For SR, our method reconstructs fine-grained textures and sharp details that are highly consistent with the ground truth (GT), in regions with complex textures and high-frequency details. This highlights the superiority of our approach in generating realistic and visually pleasing results. For LLIE, our method achieves significant improvements in both visual clarity and natural color restoration, as shown at the bottom of Figure~\ref{fig:vis_comp}. Compared to existing approaches, our results better preserve structural details and generate more accurate brightness adjustments, making them closely aligned with the GT. More visual comparisons are provided in \textbf{Appendix Section 14}. Furthermore, to evaluate the visual quality, a user study is conducted for the image super-resolution. Specifically, we randomly select 15 images from the test datasets. Fifteen participants were asked to rate the visual quality of each processed image on a scale from 0 (poor quality) to 10 (excellent quality). The aggregated results, as shown in Figure~\ref{fig:user_study}, reveal that existing methods often fail to fully restore image quality, leading to lower user satisfaction. In contrast, our method achieves the highest average score of 8.4, demonstrating superior visual performance and generalization capabilities.

\section{{Limitaion and Future Work}}
While our \NAMENetWork demonstrates strong performance on three representative restoration tasks, its evaluation is currently confined to these specific domains. Future work will aim to extend our method to a broader range of low-level vision tasks, such as image dehazing, debluring, denoising, and inpainting, among others, to further investigate its generalizability. Additionally, the top \( p\% \) of different images in the current framework is uniform, which may not fully account for the varying texture complexities of different images. Images with richer textures often pose greater challenges than those with fewer textures. Developing adaptive processing strategies, such as texture-aware top \( p\% \) selection mechanisms, could enhance performance by dynamically allocating computational resources based on image characteristics. These adaptations will be a key focus of our future research.

\section{Conclusion}

In this paper, we proposed \NAMENetWork, a novel framework for image restoration that achieves a balance between high restoration quality and computational efficiency. Leveraging the statical characteristics of image degradation, a novel \NAMESSM is introduced, which enhances texture awareness and improves performance by modulating the state-space equation and focusing more attention on texture-rich regions. Additionally, the Multi-Directional Perception Block expands the receptive field in multiple directions while maintaining low computational overhead. Extensive experiments on benchmarks for super-resolution, deraining, and low-light image enhancement demonstrate that \NAMENetWork achieves state-of-the-art performance with significant improvements in efficiency, providing a robust and effective backbone for image restoration.



\clearpage
\setcounter{page}{1}

\begin{figure}[t]
\centering
\includegraphics[width=0.5\textwidth]{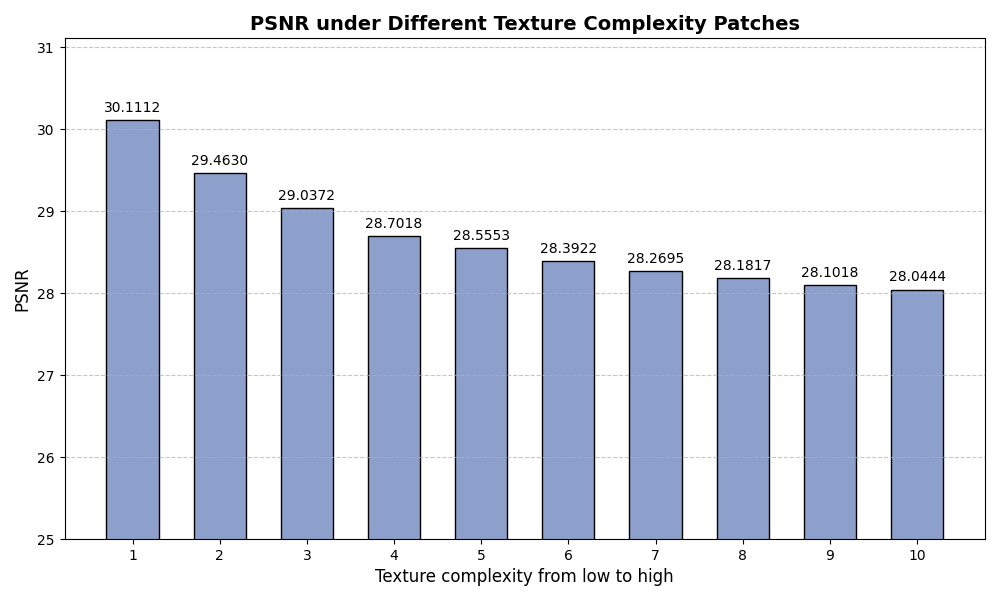}
\caption{Analysis of low-light image enhancement tasks.}
\label{fig:any_LLIE}
\end{figure}
\section{More analysis of low-light image enhancement tasks}
\label{sec:appendix_more_details_of_methods}
In Figure 1 of the main text, we explored the relationship between image super-resolution degradation and texture complexity, demonstrating that regions with richer textures suffer more severe degradation, leading to lower PSNR values. Here, we further extend this analysis to low-light image enhancement tasks. Specifically, we divide low-light images into patches and measure the texture complexity of each patch using statistical variance. We then compute the PSNR between the degraded patches and their normal-light ground truth counterparts. The results, as shown in Figure~\ref{fig:any_LLIE}, indicate that, similar to the super-resolution task, regions with richer textures in low-light image enhancement tasks also suffer more severe degradation, resulting in lower PSNR values.

\section{More details of the proposed method}
\label{sec:appendix_more_details_of_methods}

\begin{figure}[t]
\centering
\includegraphics[width=0.5\textwidth]{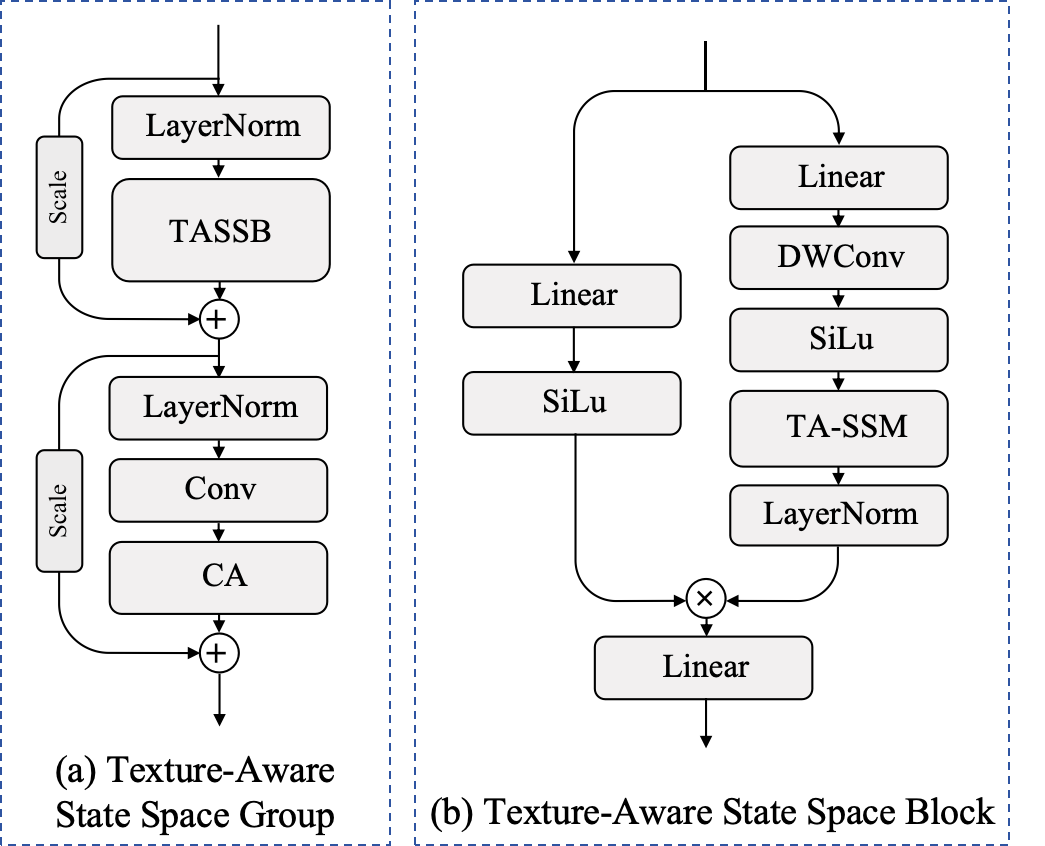}
\caption{The overall framework of the proposed texture-aware State Space Group and Texture-Aware State Space Block.}
\label{fig:framework_details}
\end{figure}

In the main text, we introduced the Texture-Aware State Space Group (TASSG) and the Texture-Aware State Space Block (TASSB) to enhance the texture-awareness capability of the State Space Model (SSM). The detailed structures of TASSG and TASSB are illustrated in Figure~\ref{fig:framework_details}. The \textbf{Texture-Aware State Space Group (TASSG)} is designed to process features at multiple scales, enabling the model to capture both global and local texture patterns. As shown in the left part of Figure~\ref{fig:framework_details}, TASSG consists of two branches. The first branch employs a convolution layer followed by a channel attention (CA) module, which can be expressed as:
\begin{equation}
\mathbf{F}_{\text{CA}} = \sigma(\mathbf{W}_2 \cdot \mathrm{ReLU}(\mathbf{W}_1 \cdot \mathrm{GAP}(\mathbf{F}))) \odot \mathbf{F},
\end{equation}
where $\mathbf{F}$ represents the input feature, $\mathrm{GAP}(\cdot)$ denotes the global average pooling operation, $\mathbf{W}_1$ and $\mathbf{W}_2$ are learnable weights, and $\sigma(\cdot)$ and $\mathrm{ReLU}(\cdot)$ are the sigmoid and ReLU activation functions, respectively. The second branch integrates LayerNorm and a stack of TASSBs to extract advanced texture features. Both branches operate on different scales and are fused by element-wise addition to form the final output.
The \textbf{Texture-Aware State Space Block (TASSB)}, as depicted in the right part of Figure~\ref{fig:framework_details}, serves as the core building block in TASSG. TASSB is carefully designed to balance efficiency and expressive power. It first employs a linear layer for feature projection:
\begin{equation}
\mathbf{F}_{\text{proj}} = \mathbf{W}_{\text{linear}} \cdot \mathbf{F}_{\text{in}},
\end{equation}
where $\mathbf{W}_{\text{linear}}$ is the weight matrix of the linear layer, and $\mathbf{F}_{\text{in}}$ is the input feature. This is followed by a depth-wise convolution (DWConv) for efficient spatial feature extraction, which can be formulated as:
\begin{equation}
\mathbf{F}_{\text{DWConv}} = \mathbf{F}_{\text{proj}} \ast \mathbf{K}_{\text{DW}},
\end{equation}
where $\ast$ denotes the convolution operation, and $\mathbf{K}_{\text{DW}}$ is the depth-wise convolution kernel. A SiLU activation function is then applied to introduce non-linearity. Subsequently, the \NAMESSM captures long-range dependencies, enhancing the model's ability to handle complex textures. The final output of TASSB is normalized using LayerNorm and passed through another linear projection layer.

In summary, the proposed TASSG and TASSB are designed to extract and process texture features at multiple scales, enhancing the overall texture-awareness capability of the model. This modular design ensures efficiency and expressive power, making the proposed method well-suited for texture-sensitive tasks.

\begin{table}[t]
\centering
\caption{Performance and complexity comparison (Params (M) and FLOPs (G)) on \underline{\textbf{low-light image enhancement}} with state-of-the-art methods.}
\label{tab:lowlightenhancement_smallversion}
\scalebox{1.0}{
\begin{tabular}{ccc|cc}
\toprule
\cellcolor[HTML]{FFFFFF}{\color[HTML]{000000} Methods} & PSNR &SSIM & Param & FLOPs \\ \midrule
RetinexNet                                             & 17.13                & 0.762                 & 0.84      & 584.47   \\
KinD                                                   & 13.29                 & 0.578                 & 8.02      & 34.99    \\
3DLUT                                                  & 18.04                 & 0.800                   & 0.59      & 7.67     \\
DRBN                                                   & 23.22                 & 0.927                 & 5.47      & 48.61    \\
LLFlow                                                 & 24.80                & 0.919                 & 17.42     & 358.40   \\
EnlightenGAN                                           & 16.57                 & 0.734                 & 114.35    & 61.01    \\
Restormer                                              & 21.41                & 0.830                  & 26.13     & 144.25   \\
LEDNet                                                 & 23.70                & 0.914                 & 7.07      & 35.92    \\
SNR-Aware                                              & 24.14                 & 0.928                 & 4.01      & 26.35    \\
LLFormer                                               & 24.03                & 0.909                 & 24.55     & 22.52    \\
\NAMENetWork-S                                             & \textbf{26.12}      & \textbf{0.939}       & 3.97     & 26.86    \\ \bottomrule
\end{tabular}
}
\end{table}

\section{More details of training process}
\label{sec:appendix_more_details_of_training}

\subsection{Image super-resolution}
Following prior works~\cite{liang2021swinir,MambaIR}, the training batch sizes for image super-resolution are set to 32, 16, and 16, respectively. We perform data augmentation by applying horizontal flips and random rotations of $90^\circ, 180^\circ$, and $270^\circ$. Additionally, we crop the original images into $64 \times 64$ patches for image SR during training.  We employ the Adam~\cite{kingma2014adam} as the optimizer for training our MambaIR with $\beta_1 = 0.9, \beta_2 = 0.999$. The initial learning rate is set at $2 \times 10^{-4}$ and is halved when the training iteration reaches specific milestones. A total of 1,000K iterations are performed to train our model. All training experiments are conducted on 8 NVIDIA V100 GPUs.

\subsection{Low-light image enhancement}
For the low-light image enhancement task, we adopt a progressive training strategy where the input patch size gradually increases from $128 \times 128$ to $196 \times 196$, $256 \times 256$, and finally $320 \times 320$. To accommodate the growing patch sizes, the batch size is progressively reduced from 64 to 48, 16, and 8, respectively. The learning rate follows the same settings as in the super-resolution task, starting at $2 \times 10^{-4}$ and halving at specific training milestones. Unlike the plain architecture used in the super-resolution task, we follow~\cite{MambaIR} to employ a UNet-based architecture to improve restoration quality in low-light conditions. Specifically, the network consists of three downsampling and three upsampling stages, with intermediate latent space processing. The downsampling and upsampling stages contain 8, 10, 10, and 12 blocks, respectively, while the latent space comprises 12 blocks. This design enables effective feature extraction and reconstruction across different spatial resolutions.

\subsection{Image deraining}
For the image deraining task, we use the same progressive training strategy as in low-light image enhancement. The input patch size gradually transitions from $128 \times 128$ to $196 \times 196$, $256 \times 256$, and finally, $320 \times 320$, while the batch size is reduced progressively from 64 to 48, 16, and 8, respectively. The learning rate is initialized at $2 \times 10^{-4}$ and halved at specific milestones during the training process. We also employ a UNet-based architecture for image deraining, which consists of three downsampling and three upsampling stages, with intermediate latent space processing. The downsampling and upsampling stages contain 8, 10, 10, and 12 blocks, respectively, while the latent space includes 12 blocks with an MLP ratio of 1.2. This architecture is designed to effectively handle complex rain streak patterns and restore high-quality visuals. 

\begin{table*}[t]
\centering
\caption{Performance comparison on Manga109 with FLOPs, Params, and Inference Time.}
\small
\begin{tabular}{lcccccccccc}
\toprule
\textbf{Metric} & {EDSR} & {RCAN} & {SAN} & {HAN} & {CSNLN} & {NLSA} & {ELAN} & {SwinIR} & {SRFormer} & \textbf{TAMambaIR-S} \\
\midrule
PSNR (dB)       & 39.10 & 39.44 & 39.32 & 39.46 & 39.37 & 39.59 & 39.93 & 39.92 & 40.07 & \textbf{40.23} \\
SSIM            & 0.9773 & 0.9786 & 0.9792 & 0.9785 & 0.9785 & 0.9789 & 0.9800 & 0.9797 & 0.9802 & \textbf{0.9806} \\
FLOPs (G)       & 166.84 & 62.75 & 64.11 & 258.82 & 481.97 & 182.82 & 30.22 & 51.33 & 62.95 & 56.88 \\
Params (M)      & 40.73 & 15.44 & 15.71 & 63.61 & 6.21 & 41.80 & 11.48 & 11.75 & 10.40 & 12.19 \\
Inference Time (ms) & 18.26 & 14.19 & 14.46 & 95.04 & 124.37 & 30.12 & 50.51 & 53.63 & 70.18 & 59.30 \\
\bottomrule
\end{tabular}
\label{tab:manga109}
\end{table*}

\section{More details of datasets}
\label{sec:appendix_more_details_of_datasets}
We conduct experiments on several widely used and publicly available datasets to ensure fair and comprehensive evaluations. For the image super-resolution task, DIV2K~\cite{timofte2017ntire} and Flickr2K~\cite{lim2017enhanced} are utilized to train the network, while Set5~\cite{bevilacqua2012low}, Set14~\cite{zeyde2012single}, B100~\cite{martin2001database}, and Manga109~\cite{matsui2017sketch} are adopted for evaluation. For the image deraining task, Rain200L~\cite{yang2017deep} and Rain200H~\cite{yang2017deep} are used for evaluation. These datasets contain light and heavy rain conditions, respectively, and include 1,800 rainy images for training and 200 rainy images for testing. For the low-light image enhancement task, we follow the setup in~\cite{cai2023retinexformer} and validate our approach on the synthetic version of the LOL-V2 dataset~\cite{LOLv2}, which consists of 900 low-light and normal-light image pairs for training and 100 pairs for testing.

\section{More details of compared state-of-the-art methods}
\label{sec:appendix_more_details_of_compared_methods}
We evaluate our method on three classic and widely studied image restoration tasks, including image super-resolution, image deraining, and low-light image enhancement. For each task, we select the most representative datasets and compare our approach against existing state-of-the-art methods to demonstrate its effectiveness and robustness. \textbf{{Image super-resolution}}: we compare our method against thirteen state-of-the-art methods, including: RCAN~\cite{zhang2018rcan}, SAN~\cite{dai2019second}, ClassSR~\cite{Kong_2021_CVPR}, IPT~\cite{chen2021pre}, CSNLN~\cite{mei2020image}, EDT-B~\cite{li2021efficient}, EDSR~\cite{lim2017enhanced}, RDN~\cite{zhang2018residual}, HAN~\cite{niu2020single}, SwinIR~\cite{liang2021swinir}, SRFormer~\cite{zhou2023srformer}, MambaIR~\cite{MambaIR}. \textbf{{Image deraining}}: we compare our method against fifteen state-of-the-art methods, including: DDN~\cite{fu2017removing}, RESCAN~\cite{li2018recurrent}, PReNet~\cite{ren2019progressive}, MSPFN~\cite{jiang2020multi}, RCDNet~\cite{wang2020model} MPRNet~\cite{zamir2021multi}, SwinIR~\cite{liang2021swinir}, DualGCN~\cite{fu2021rain}, SPDNet~\cite{yi2021structure}, Uformer~\cite{wang2022uformer}, Restormer~\cite{zamir2022restormer}, IDT~\cite{xiao2022image}, DLINet~\cite{DLINet}, DRSformer~\cite{Chen_2023_CVPR} and MambaIR~\cite{MambaIR}. \textbf{{Low-light image enhancement}}: we compare our method against fifteen state-of-the-art methods, including: RetinexNet~\cite{RetinexNet}, KinD~\cite{KinD}, ZeroDCE~\cite{Zero-DCE}, 3DLUT~\cite{3DLUT}, DRBN~\cite{DRBN}, RUAS~\cite{RUAS}, LLFlow~\cite{LLFlow}, EnlightenGAN~\cite{EnGAN}, Restormer~\cite{Restormer}, LEDNet~\cite{LEDNet}, SNR-Aware~\cite{SNR-Aware}, LLFormer~\cite{LLFormer}, RetinexFormer~\cite{RetinexFormer}, CIDNet\cite{CIDNet} and MambaIR~\cite{MambaIR}.

\begin{table*}[t]
\centering
\caption{Performance comparison on Manga109 with PSNR and FLOPs under different Top \( p\% \).}
\begin{tabular}{c|ccccccc}
\toprule
Top \( p\% \) & 20\%   & 3020\%    & 40\%    & 50\%   & 60\%    & 70\%    & 80\%    \\ \midrule
PSNR                                                 & 40.05 & 40.14  & 40.17  & 40.23 & 40.26  & 40.27  & 40.28  \\
FLOPs                                                & 0.085 & 0.1278 & 0.1705 & 0.213 & 0.2557 & 0.2983 & 0.3409 \\ \bottomrule
\end{tabular}
\label{tab:abs_topk}
\end{table*}

\section{More ablation studies}
\label{sec:appendix_more_ab_study}
In this section, we provide additional ablation studies, including the texture measurement methods and the setting of the top \( p\% \). Specifically, we conduct these experiments on the \NAMENetWork-S model using the Manga109 dataset. For texture measurement methods, we consider the statistical variance as it effectively reflects the dispersion within local patches and aligns well with image textures. Additionally, we replaced the statistical variance with information entropy as the texture measurement method and observed a 0.05 dB drop in PSNR. Furthermore, the higher computational complexity of information entropy led to slower processing speeds, while the statistical variance proved to be computationally efficient and yielded better performance in this experiment.

Furthermore, we also investigate the impact of the top \( p\% \) selection. Specifically, we varied \( p \) from 0.2 to 0.8 and conducted experiments to record the corresponding PSNR values and the computational cost (measured in FLOPs) of the \NAMESSM module. The detailed results are presented in Table~\ref{tab:abs_topk}. From the results, we observe a significant increase in PSNR as \( p \) grows from 0.2 to 0.5, accompanied by a steady rise in FLOPs. However, beyond \( p = 0.5 \), the PSNR improvement slows down, while FLOPs continue to increase substantially. Based on this observation, we conclude that \( p = 50\% \) strikes a good balance between performance and computational efficiency. Therefore, in this work, we empirically set the top \( p\% \) to 50.

\section{More comparison experiment}
\label{sec:appendix_more_compared_exp}

\subsection{Low-Light Image Enhancement}
To further demonstrate the efficiency of our proposed method, we conduct a performance and complexity comparison on the low-light image enhancement task, as shown in Table~\ref{tab:lowlightenhancement_smallversion}. Specifically, we design a small version of our model to evaluate its effectiveness under resource-constrained settings by changing the block and middle channel feature number. The results highlight the superior performance of our method in terms of both image quality and computational efficiency. Our small model achieves a PSNR of 26.12 dB and an SSIM of 0.939, outperforming state-of-the-art methods, including LLFlow (24.80/0.919) and Restormer (21.41/0.830). At the same time, our model exhibits remarkable efficiency, requiring only 3.97M parameters and 26.86G FLOPs, which is significantly lower than computationally intensive methods like EnlightenGAN (114.35M, 61.01G) and LLFlow (17.42M, 358.40G). These results demonstrate that our small model not only achieves state-of-the-art performance but also maintains exceptional computational efficiency, making it a highly practical solution for image enhancement tasks.

\subsection{Image Super-Resolution}

Table~\ref{tab:manga109} provides a detailed comparison of the small version of our method (TAMambaIR-S) against previous state-of-the-art (SOTA) approaches on the Manga109 dataset. Our method achieves the highest PSNR of 40.23 dB and SSIM of 0.9806. Furthermore, TAMambaIR-S demonstrates improved efficiency, with significantly lower FLOPs (56.88G) compared to other SOTA methods, such as SRFormer (62.95G) and NLSA (182.82G). In terms of inference time, TAMambaIR-S achieves a competitive runtime of 59.30 ms, outperforming computationally heavy models like HAN (95.04 ms) and CSNLN (124.37 ms). This demonstrates the superior balance of our method in achieving state-of-the-art performance, computational efficiency, and faster runtime.

\begin{figure*}[t]
\centering
\includegraphics[width=1.0\textwidth]{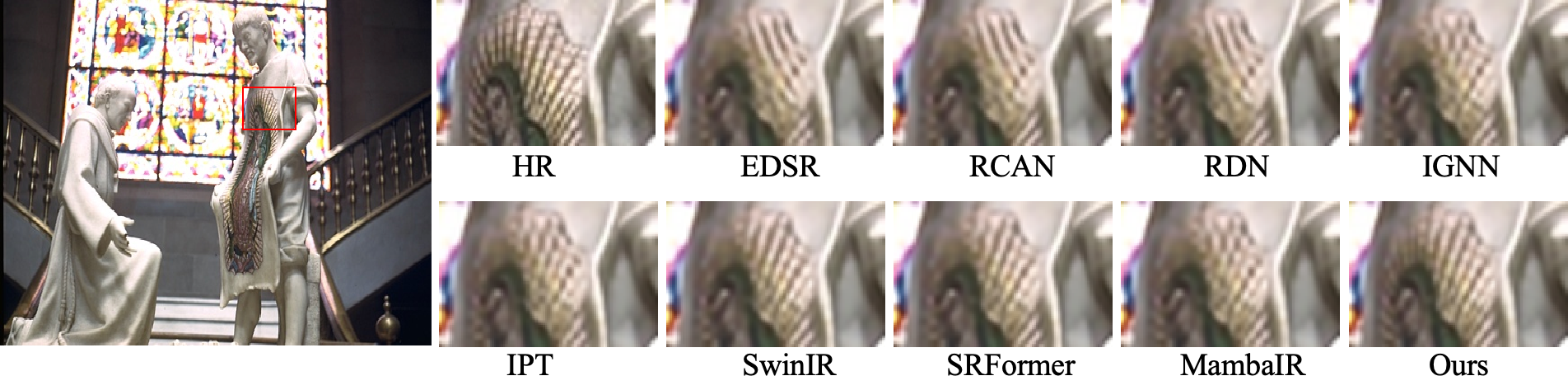}
\caption{Additional visual comparison on image super-resolution. Please zoom in for a better view.}
\label{fig:vis_comp_SR2}
\end{figure*}

\begin{figure*}[t]
\centering
\includegraphics[width=1.0\textwidth]{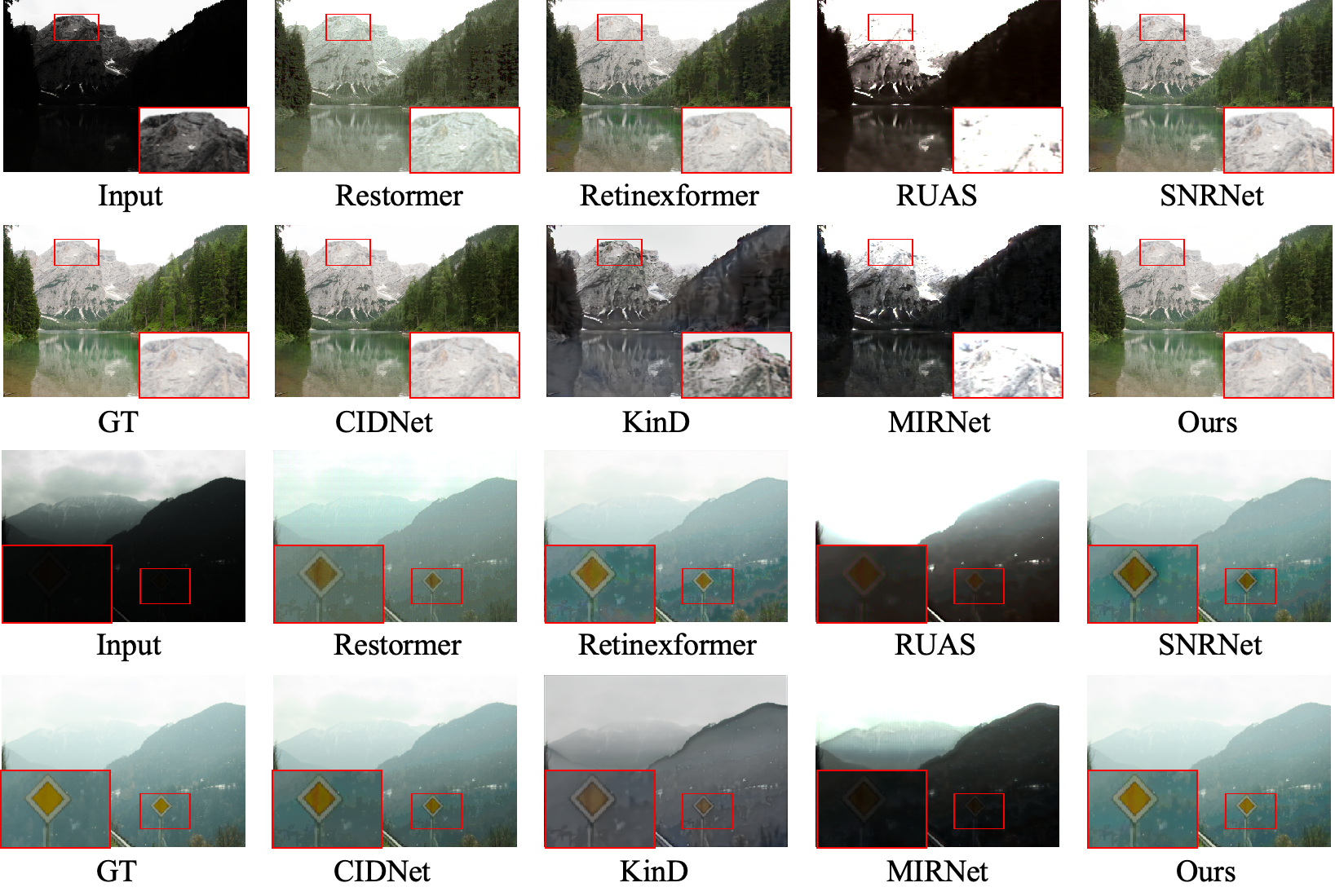}
\caption{Additional visual comparison on low-light image enhancement. Please zoom in for a better view.}
\label{fig:Vis_Comp_LLIE2}
\end{figure*}

\section{More visual comparisons}
\label{sec:appendix_more_vis_comp}

In this section, we provide additional visual comparisons to further validate the effectiveness of our method. Specifically, we showcase more results for both super-resolution (SR) and low-light image enhancement (LLIE) tasks. These visualizations demonstrate the superiority of our approach in handling challenging scenarios, such as complex textures, fine details, and extreme lighting conditions. For each task, we compare our results with those of state-of-the-art methods to highlight the qualitative improvements achieved by our method, as shown in Figure~\ref{fig:vis_comp_SR2} and~\ref{fig:Vis_Comp_LLIE2}.
\clearpage
\bibliographystyle{named}
\bibliography{ijcai25}

\begin{thebibliography}{}

\bibitem[\protect\citeauthoryear{Bevilacqua \bgroup \em et al.\egroup }{2012}]{bevilacqua2012low}
Marco Bevilacqua, Aline Roumy, Christine Guillemot, and Marie~Line Alberi-Morel.
\newblock Low-complexity single-image super-resolution based on nonnegative neighbor embedding.
\newblock 2012.

\bibitem[\protect\citeauthoryear{Cai \bgroup \em et al.\egroup }{2023a}]{RetinexFormer}
Yuanhao Cai, Hao Bian, Jing Lin, Haoqian Wang, Radu Timofte, and Yulun Zhang.
\newblock Retinexformer: One-stage retinex-based transformer for low-light image enhancement.
\newblock In {\em Proceedings of the IEEE/CVF International Conference on Computer Vision (ICCV)}, pages 12504--12513, October 2023.

\bibitem[\protect\citeauthoryear{Cai \bgroup \em et al.\egroup }{2023b}]{cai2023retinexformer}
Yuanhao Cai, Hao Bian, Jing Lin, Haoqian Wang, Radu Timofte, and Yulun Zhang.
\newblock Retinexformer: One-stage retinex-based transformer for low-light image enhancement.
\newblock In {\em Proceedings of the IEEE/CVF International Conference on Computer Vision}, pages 12504--12513, 2023.

\bibitem[\protect\citeauthoryear{Chen \bgroup \em et al.\egroup }{2021}]{chen2021pre}
Hanting Chen, Yunhe Wang, Tianyu Guo, Chang Xu, Yiping Deng, Zhenhua Liu, Siwei Ma, Chunjing Xu, Chao Xu, and Wen Gao.
\newblock Pre-trained image processing transformer.
\newblock In {\em CVPR}, pages 12299--12310, 2021.

\bibitem[\protect\citeauthoryear{Chen \bgroup \em et al.\egroup }{2023a}]{Chen_2023_CVPR}
Xiang Chen, Hao Li, Mingqiang Li, and Jinshan Pan.
\newblock Learning a sparse transformer network for effective image deraining.
\newblock In {\em Proceedings of the IEEE/CVF Conference on Computer Vision and Pattern Recognition (CVPR)}, pages 5896--5905, June 2023.

\bibitem[\protect\citeauthoryear{Chen \bgroup \em et al.\egroup }{2023b}]{DRSformer}
Xiang Chen, Hao Li, Mingqiang Li, and Jinshan Pan.
\newblock Learning a sparse transformer network for effective image deraining.
\newblock In {\em Proceedings of the IEEE/CVF Conference on Computer Vision and Pattern Recognition (CVPR)}, pages 5896--5905, June 2023.

\bibitem[\protect\citeauthoryear{Chen \bgroup \em et al.\egroup }{2024}]{chen2024changemamba}
Hongruixuan Chen, Jian Song, Chengxi Han, Junshi Xia, and Naoto Yokoya.
\newblock Changemamba: Remote sensing change detection with spatio-temporal state space model.
\newblock {\em arXiv preprint arXiv:2404.03425}, 2024.

\bibitem[\protect\citeauthoryear{Cui \bgroup \em et al.\egroup }{2023}]{focal}
Yuning Cui, Wenqi Ren, Xiaochun Cao, and Alois Knoll.
\newblock Focal network for image restoration.
\newblock In {\em Proceedings of the IEEE/CVF international conference on computer vision}, pages 13001--13011, 2023.

\bibitem[\protect\citeauthoryear{Dai \bgroup \em et al.\egroup }{2019}]{dai2019second}
Tao Dai, Jianrui Cai, Yongbing Zhang, Shu-Tao Xia, and Lei Zhang.
\newblock Second-order attention network for single image super-resolution.
\newblock In {\em Proceedings of the IEEE/CVF conference on computer vision and pattern recognition}, pages 11065--11074, 2019.

\bibitem[\protect\citeauthoryear{Dong \bgroup \em et al.\egroup }{2015}]{dong2015image}
Chao Dong, Chen~Change Loy, Kaiming He, and Xiaoou Tang.
\newblock Image super-resolution using deep convolutional networks.
\newblock {\em IEEE transactions on pattern analysis and machine intelligence}, 38(2):295--307, 2015.

\bibitem[\protect\citeauthoryear{Dosovitskiy \bgroup \em et al.\egroup }{2020}]{a:31}
Alexey Dosovitskiy, Lucas Beyer, Alexander Kolesnikov, Dirk Weissenborn, Xiaohua Zhai, Thomas Unterthiner, Mostafa Dehghani, Matthias Minderer, Georg Heigold, Sylvain Gelly, et~al.
\newblock An image is worth 16x16 words: Transformers for image recognition at scale.
\newblock {\em arXiv preprint arXiv:2010.11929}, 2020.

\bibitem[\protect\citeauthoryear{Fang \bgroup \em et al.\egroup }{2022}]{fang2022hybrid}
Jinsheng Fang, Hanjiang Lin, Xinyu Chen, and Kun Zeng.
\newblock A hybrid network of cnn and transformer for lightweight image super-resolution.
\newblock In {\em Proceedings of the IEEE/CVF conference on computer vision and pattern recognition}, pages 1103--1112, 2022.

\bibitem[\protect\citeauthoryear{Feng \bgroup \em et al.\egroup }{2024}]{CIDNet}
Yixu Feng, Cheng Zhang, Pei Wang, Peng Wu, Qingsen Yan, and Yanning Zhang.
\newblock You only need one color space: An efficient network for low-light image enhancement, 2024.

\bibitem[\protect\citeauthoryear{Fu \bgroup \em et al.\egroup }{2017}]{fu2017removing}
Xueyang Fu, Jiabin Huang, Delu Zeng, Yue Huang, Xinghao Ding, and John Paisley.
\newblock Removing rain from single images via a deep detail network.
\newblock In {\em CVPR}, pages 3855--3863, 2017.

\bibitem[\protect\citeauthoryear{Fu \bgroup \em et al.\egroup }{2021}]{fu2021rain}
Xueyang Fu, Qi~Qi, Zheng-Jun Zha, Yurui Zhu, and Xinghao Ding.
\newblock Rain streak removal via dual graph convolutional network.
\newblock In {\em AAAI}, pages 1352--1360, 2021.

\bibitem[\protect\citeauthoryear{Fu \bgroup \em et al.\egroup }{2024}]{fu2024ssumamba}
Guanyiman Fu, Fengchao Xiong, Jianfeng Lu, Jun Zhou, and Yuntao Qian.
\newblock Ssumamba: Spatial-spectral selective state space model for hyperspectral image denoising.
\newblock {\em arXiv preprint arXiv:2405.01726}, 2024.

\bibitem[\protect\citeauthoryear{Guo \bgroup \em et al.\egroup }{2020}]{Zero-DCE}
Chunle~Guo Guo, Chongyi Li, Jichang Guo, Chen~Change Loy, Junhui Hou, Sam Kwong, and Runmin Cong.
\newblock Zero-reference deep curve estimation for low-light image enhancement.
\newblock In {\em Proceedings of the IEEE conference on computer vision and pattern recognition (CVPR)}, pages 1780--1789, June 2020.

\bibitem[\protect\citeauthoryear{Guo \bgroup \em et al.\egroup }{2024a}]{MambaIR}
Hang Guo, Jinmin Li, Tao Dai, Zhihao Ouyang, Xudong Ren, and Shu-Tao Xia.
\newblock Mambair: A simple baseline for image restoration with state-space model.
\newblock In {\em ECCV}, 2024.

\bibitem[\protect\citeauthoryear{Guo \bgroup \em et al.\egroup }{2024b}]{c:8}
Hang Guo, Jinmin Li, Tao Dai, Zhihao Ouyang, Xudong Ren, and Shu-Tao Xia.
\newblock Mambair: A simple baseline for image restoration with state-space model.
\newblock {\em arXiv preprint arXiv:2402.15648}, 2024.

\bibitem[\protect\citeauthoryear{He \bgroup \em et al.\egroup }{2024}]{he2024latent}
Yuhong He, Long Peng, Lu~Wang, and Jun Cheng.
\newblock Latent degradation representation constraint for single image deraining.
\newblock In {\em ICASSP 2024-2024 IEEE International Conference on Acoustics, Speech and Signal Processing (ICASSP)}, pages 3155--3159. IEEE, 2024.

\bibitem[\protect\citeauthoryear{Jeong \bgroup \em et al.\egroup }{2025}]{jeong2025accelerating}
Jinho Jeong, Jinwoo Kim, Younghyun Jo, and Seon~Joo Kim.
\newblock Accelerating image super-resolution networks with pixel-level classification.
\newblock In {\em European Conference on Computer Vision}, pages 236--251. Springer, 2025.

\bibitem[\protect\citeauthoryear{Jiang \bgroup \em et al.\egroup }{2020}]{jiang2020multi}
Kui Jiang, Zhongyuan Wang, Peng Yi, Chen Chen, Baojin Huang, Yimin Luo, Jiayi Ma, and Junjun Jiang.
\newblock Multi-scale progressive fusion network for single image deraining.
\newblock In {\em CVPR}, pages 8346--8355, 2020.

\bibitem[\protect\citeauthoryear{Jiang \bgroup \em et al.\egroup }{2021}]{EnGAN}
Yifan Jiang, Xinyu Gong, Ding Liu, Yu~Cheng, Chen Fang, Xiaohui Shen, Jianchao Yang, Pan Zhou, and Zhangyang Wang.
\newblock Enlightengan: Deep light enhancement without paired supervision.
\newblock {\em IEEE Transactions on Image Processing}, 30:2340--2349, 2021.

\bibitem[\protect\citeauthoryear{Kalman}{1960}]{a:25}
Rudolph~Emil Kalman.
\newblock A new approach to linear filtering and prediction problems.
\newblock 1960.

\bibitem[\protect\citeauthoryear{Kingma and Ba}{2014}]{kingma2014adam}
Diederik~P Kingma and Jimmy Ba.
\newblock Adam: A method for stochastic optimization.
\newblock {\em arXiv preprint arXiv:1412.6980}, 2014.

\bibitem[\protect\citeauthoryear{Kong \bgroup \em et al.\egroup }{2021}]{Kong_2021_CVPR}
Xiangtao Kong, Hengyuan Zhao, Yu~Qiao, and Chao Dong.
\newblock Classsr: A general framework to accelerate super-resolution networks by data characteristic.
\newblock In {\em Proceedings of the IEEE/CVF Conference on Computer Vision and Pattern Recognition (CVPR)}, pages 12016--12025, June 2021.

\bibitem[\protect\citeauthoryear{Li \bgroup \em et al.\egroup }{2018}]{li2018recurrent}
Xia Li, Jianlong Wu, Zhouchen Lin, Hong Liu, and Hongbin Zha.
\newblock Recurrent squeeze-and-excitation context aggregation net for single image deraining.
\newblock In {\em ECCV}, pages 254--269, 2018.

\bibitem[\protect\citeauthoryear{Li \bgroup \em et al.\egroup }{2021}]{li2021efficient}
Wenbo Li, Xin Lu, Shengju Qian, Jiangbo Lu, Xiangyu Zhang, and Jiaya Jia.
\newblock On efficient transformer and image pre-training for low-level vision.
\newblock {\em arXiv preprint arXiv:2112.10175}, 2021.

\bibitem[\protect\citeauthoryear{Li \bgroup \em et al.\egroup }{2022}]{2022LLE}
Chongyi Li, Chunle Guo, Linghao Han, Jun Jiang, Ming-Ming Cheng, Jinwei Gu, and Chen~Change Loy.
\newblock Low-light image and video enhancement using deep learning: A survey.
\newblock {\em IEEE Transactions on Pattern Analysis and Machine Intelligence}, 44(12):9396--9416, 2022.

\bibitem[\protect\citeauthoryear{Li \bgroup \em et al.\egroup }{2023}]{DLINet}
Wencheng Li, Gang Chen, and Yi~Chang.
\newblock An efficient single image de-raining model with decoupled deep networks.
\newblock {\em IEEE Transactions on Image Processing}, 33:69--81, 2023.

\bibitem[\protect\citeauthoryear{Li \bgroup \em et al.\egroup }{2024}]{a:36}
Dong Li, Yidi Liu, Xueyang Fu, Senyan Xu, and Zheng-Jun Zha.
\newblock Fouriermamba: Fourier learning integration with state space models for image deraining.
\newblock {\em arXiv preprint arXiv:2405.19450}, 2024.

\bibitem[\protect\citeauthoryear{Liang \bgroup \em et al.\egroup }{2021a}]{a:22}
Jingyun Liang, Jiezhang Cao, Guolei Sun, Kai Zhang, Luc Van~Gool, and Radu Timofte.
\newblock Swinir: Image restoration using swin transformer.
\newblock In {\em Proceedings of the IEEE/CVF international conference on computer vision}, pages 1833--1844, 2021.

\bibitem[\protect\citeauthoryear{Liang \bgroup \em et al.\egroup }{2021b}]{liang2021swinir}
Jingyun Liang, Jiezhang Cao, Guolei Sun, Kai Zhang, Luc Van~Gool, and Radu Timofte.
\newblock Swinir: Image restoration using swin transformer.
\newblock In {\em Proceedings of the IEEE/CVF international conference on computer vision}, pages 1833--1844, 2021.

\bibitem[\protect\citeauthoryear{Lim \bgroup \em et al.\egroup }{2017}]{lim2017enhanced}
Bee Lim, Sanghyun Son, Heewon Kim, Seungjun Nah, and Kyoung Mu~Lee.
\newblock Enhanced deep residual networks for single image super-resolution.
\newblock In {\em Proceedings of the IEEE conference on computer vision and pattern recognition workshops}, pages 136--144, 2017.

\bibitem[\protect\citeauthoryear{Martin \bgroup \em et al.\egroup }{2001}]{martin2001database}
David Martin, Charless Fowlkes, Doron Tal, and Jitendra Malik.
\newblock A database of human segmented natural images and its application to evaluating segmentation algorithms and measuring ecological statistics.
\newblock In {\em Proceedings Eighth IEEE International Conference on Computer Vision. ICCV 2001}, volume~2, pages 416--423. IEEE, 2001.

\bibitem[\protect\citeauthoryear{Matsui \bgroup \em et al.\egroup }{2017}]{matsui2017sketch}
Yusuke Matsui, Kota Ito, Yuji Aramaki, Azuma Fujimoto, Toru Ogawa, Toshihiko Yamasaki, and Kiyoharu Aizawa.
\newblock Sketch-based manga retrieval using manga109 dataset.
\newblock {\em Multimedia Tools and Applications}, 76:21811--21838, 2017.

\bibitem[\protect\citeauthoryear{Mei \bgroup \em et al.\egroup }{2020}]{mei2020image}
Yiqun Mei, Yuchen Fan, Yuqian Zhou, Lichao Huang, Thomas~S Huang, and Honghui Shi.
\newblock Image super-resolution with cross-scale non-local attention and exhaustive self-exemplars mining.
\newblock In {\em Proceedings of the IEEE/CVF conference on computer vision and pattern recognition}, pages 5690--5699, 2020.

\bibitem[\protect\citeauthoryear{Niu \bgroup \em et al.\egroup }{2020}]{niu2020single}
Ben Niu, Weilei Wen, Wenqi Ren, Xiangde Zhang, Lianping Yang, Shuzhen Wang, Kaihao Zhang, Xiaochun Cao, and Haifeng Shen.
\newblock Single image super-resolution via a holistic attention network.
\newblock In {\em Computer Vision--ECCV 2020: 16th European Conference, Glasgow, UK, August 23--28, 2020, Proceedings, Part XII 16}, pages 191--207. Springer, 2020.

\bibitem[\protect\citeauthoryear{Patro and Agneeswaran}{2024}]{patro2024simba}
Badri~N Patro and Vijay~S Agneeswaran.
\newblock Simba: Simplified mamba-based architecture for vision and multivariate time series.
\newblock {\em arXiv preprint arXiv:2403.15360}, 2024.

\bibitem[\protect\citeauthoryear{Peng \bgroup \em et al.\egroup }{2020}]{peng2020cumulative}
Long Peng, Aiwen Jiang, Qiaosi Yi, and Mingwen Wang.
\newblock Cumulative rain density sensing network for single image derain.
\newblock {\em IEEE Signal Processing Letters}, 27:406--410, 2020.

\bibitem[\protect\citeauthoryear{Peng \bgroup \em et al.\egroup }{2021}]{peng2021ensemble}
Long Peng, Aiwen Jiang, Haoran Wei, Bo~Liu, and Mingwen Wang.
\newblock Ensemble single image deraining network via progressive structural boosting constraints.
\newblock {\em Signal Processing: Image Communication}, 99:116460, 2021.

\bibitem[\protect\citeauthoryear{Peng \bgroup \em et al.\egroup }{2024a}]{peng2024efficient}
Long Peng, Yang Cao, Renjing Pei, Wenbo Li, Jiaming Guo, Xueyang Fu, Yang Wang, and Zheng-Jun Zha.
\newblock Efficient real-world image super-resolution via adaptive directional gradient convolution.
\newblock {\em arXiv preprint arXiv:2405.07023}, 2024.

\bibitem[\protect\citeauthoryear{Peng \bgroup \em et al.\egroup }{2024b}]{peng2024lightweight}
Long Peng, Yang Cao, Yuejin Sun, and Yang Wang.
\newblock Lightweight adaptive feature de-drifting for compressed image classification.
\newblock {\em IEEE Transactions on Multimedia}, 2024.

\bibitem[\protect\citeauthoryear{Peng \bgroup \em et al.\egroup }{2024c}]{peng2024unveiling}
Long Peng, Wenbo Li, Jiaming Guo, Xin Di, Haoze Sun, Yong Li, Renjing Pei, Yang Wang, Yang Cao, and Zheng-Jun Zha.
\newblock Unveiling hidden details: A raw data-enhanced paradigm for real-world super-resolution.
\newblock {\em arXiv preprint arXiv:2411.10798}, 2024.

\bibitem[\protect\citeauthoryear{Peng \bgroup \em et al.\egroup }{2024d}]{peng2024towards}
Long Peng, Wenbo Li, Renjing Pei, Jingjing Ren, Yang Wang, Yang Cao, and Zheng-Jun Zha.
\newblock Towards realistic data generation for real-world super-resolution.
\newblock {\em arXiv preprint arXiv:2406.07255}, 2024.

\bibitem[\protect\citeauthoryear{Qiao \bgroup \em et al.\egroup }{2024}]{a:32}
Yanyuan Qiao, Zheng Yu, Longteng Guo, Sihan Chen, Zijia Zhao, Mingzhen Sun, Qi~Wu, and Jing Liu.
\newblock Vl-mamba: Exploring state space models for multimodal learning.
\newblock {\em arXiv preprint arXiv:2403.13600}, 2024.

\bibitem[\protect\citeauthoryear{Ren \bgroup \em et al.\egroup }{2019}]{ren2019progressive}
Dongwei Ren, Wangmeng Zuo, Qinghua Hu, Pengfei Zhu, and Deyu Meng.
\newblock Progressive image deraining networks: A better and simpler baseline.
\newblock In {\em CVPR}, pages 3937--3946, 2019.

\bibitem[\protect\citeauthoryear{Rim \bgroup \em et al.\egroup }{2020}]{RealBlur}
Jaesung Rim, Haeyun Lee, Jucheol Won, and Sunghyun Cho.
\newblock Real-world blur dataset for learning and benchmarking deblurring algorithms.
\newblock In {\em Proceedings of the European Conference on Computer Vision (ECCV)}, 2020.

\bibitem[\protect\citeauthoryear{Risheng \bgroup \em et al.\egroup }{2021}]{RUAS}
Liu Risheng, Ma~Long, Zhang Jiaao, Fan Xin, and Luo Zhongxuan.
\newblock Retinex-inspired unrolling with cooperative prior architecture search for low-light image enhancement.
\newblock In {\em Proceedings of the IEEE Conference on Computer Vision and Pattern Recognition}, 2021.

\bibitem[\protect\citeauthoryear{Sun \bgroup \em et al.\egroup }{2023}]{sun2023spatially}
Long Sun, Jiangxin Dong, Jinhui Tang, and Jinshan Pan.
\newblock Spatially-adaptive feature modulation for efficient image super-resolution.
\newblock In {\em Proceedings of the IEEE/CVF International Conference on Computer Vision}, pages 13190--13199, 2023.

\bibitem[\protect\citeauthoryear{Tang \bgroup \em et al.\egroup }{2024}]{a:33}
Yujin Tang, Peijie Dong, Zhenheng Tang, Xiaowen Chu, and Junwei Liang.
\newblock Vmrnn: Integrating vision mamba and lstm for efficient and accurate spatiotemporal forecasting.
\newblock In {\em Proceedings of the IEEE/CVF Conference on Computer Vision and Pattern Recognition}, pages 5663--5673, 2024.

\bibitem[\protect\citeauthoryear{Timofte \bgroup \em et al.\egroup }{2017}]{timofte2017ntire}
Radu Timofte, Eirikur Agustsson, Luc Van~Gool, Ming-Hsuan Yang, and Lei Zhang.
\newblock Ntire 2017 challenge on single image super-resolution: Methods and results.
\newblock In {\em Proceedings of the IEEE conference on computer vision and pattern recognition workshops}, pages 114--125, 2017.

\bibitem[\protect\citeauthoryear{Tsai and Huang}{1984}]{a:37}
Roger~Y Tsai and Thomas~S Huang.
\newblock Multiframe image restoration and registration.
\newblock {\em Multiframe image restoration and registration}, 1:317--339, 1984.

\bibitem[\protect\citeauthoryear{Wang \bgroup \em et al.\egroup }{2018}]{wang2018esrgan}
Xintao Wang, Ke~Yu, Shixiang Wu, Jinjin Gu, Yihao Liu, Chao Dong, Yu~Qiao, and Chen~Change Loy.
\newblock Esrgan: Enhanced super-resolution generative adversarial networks.
\newblock In {\em The European Conference on Computer Vision Workshops (ECCVW)}, September 2018.

\bibitem[\protect\citeauthoryear{Wang \bgroup \em et al.\egroup }{2020}]{wang2020model}
Hong Wang, Qi~Xie, Qian Zhao, and Deyu Meng.
\newblock A model-driven deep neural network for single image rain removal.
\newblock In {\em CVPR}, pages 3103--3112, 2020.

\bibitem[\protect\citeauthoryear{Wang \bgroup \em et al.\egroup }{2021}]{LLFlow}
Yufei Wang, Renjie Wan, Wenhan Yang, Haoliang Li, Lap-Pui Chau, and Alex~C Kot.
\newblock Low-light image enhancement with normalizing flow.
\newblock {\em arXiv preprint arXiv:2109.05923}, 2021.

\bibitem[\protect\citeauthoryear{Wang \bgroup \em et al.\egroup }{2022}]{wang2022uformer}
Zhendong Wang, Xiaodong Cun, Jianmin Bao, Wengang Zhou, Jianzhuang Liu, and Houqiang Li.
\newblock Uformer: A general u-shaped transformer for image restoration.
\newblock In {\em Proceedings of the IEEE/CVF Conference on Computer Vision and Pattern Recognition (CVPR)}, pages 17683--17693, June 2022.

\bibitem[\protect\citeauthoryear{Wang \bgroup \em et al.\egroup }{2023a}]{wang2023brightness}
Haodian Wang, Long Peng, Yuejin Sun, Zengyu Wan, Yang Wang, and Yang Cao.
\newblock Brightness perceiving for recursive low-light image enhancement.
\newblock {\em IEEE Transactions on Artificial Intelligence}, 2023.

\bibitem[\protect\citeauthoryear{Wang \bgroup \em et al.\egroup }{2023b}]{LLFormer}
Tao Wang, Kaihao Zhang, Tianrun Shen, Wenhan Luo, Bjorn Stenger, and Tong Lu.
\newblock Ultra-high-definition low-light image enhancement: A benchmark and transformer-based method.
\newblock In {\em Proceedings of the AAAI Conference on Artificial Intelligence}, volume~37, pages 2654--2662, 2023.

\bibitem[\protect\citeauthoryear{Wang \bgroup \em et al.\egroup }{2023c}]{wang2023decoupling}
Yang Wang, Long Peng, Liang Li, Yang Cao, and Zheng-Jun Zha.
\newblock Decoupling-and-aggregating for image exposure correction.
\newblock In {\em Proceedings of the IEEE/CVF Conference on Computer Vision and Pattern Recognition}, pages 18115--18124, 2023.

\bibitem[\protect\citeauthoryear{Wang \bgroup \em et al.\egroup }{2024}]{a:34}
Ziyang Wang, Jian-Qing Zheng, Yichi Zhang, Ge~Cui, and Lei Li.
\newblock Mamba-unet: Unet-like pure visual mamba for medical image segmentation.
\newblock {\em arXiv preprint arXiv:2402.05079}, 2024.

\bibitem[\protect\citeauthoryear{Wei \bgroup \em et al.\egroup }{2018}]{RetinexNet}
Chen Wei, Wenjing Wang, Wenhan Yang, and Jiaying Liu.
\newblock Deep retinex decomposition for low-light enhancement.
\newblock 2018.

\bibitem[\protect\citeauthoryear{Xiao \bgroup \em et al.\egroup }{2022}]{xiao2022image}
Jie Xiao, Xueyang Fu, Aiping Liu, Feng Wu, and Zheng-Jun Zha.
\newblock Image de-raining transformer.
\newblock {\em IEEE TPAMI}, 2022.

\bibitem[\protect\citeauthoryear{Xu \bgroup \em et al.\egroup }{2022}]{SNR-Aware}
Xiaogang Xu, Ruixing Wang, Chi-Wing Fu, and Jiaya Jia.
\newblock Snr-aware low-light image enhancement.
\newblock In {\em 2022 IEEE/CVF Conference on Computer Vision and Pattern Recognition (CVPR)}, pages 17693--17703, 2022.

\bibitem[\protect\citeauthoryear{Yang \bgroup \em et al.\egroup }{2017}]{yang2017deep}
Wenhan Yang, Robby~T Tan, Jiashi Feng, Jiaying Liu, Zongming Guo, and Shuicheng Yan.
\newblock Deep joint rain detection and removal from a single image.
\newblock In {\em CVPR}, pages 1357--1366, 2017.

\bibitem[\protect\citeauthoryear{Yang \bgroup \em et al.\egroup }{2020}]{DRBN}
Wenhan Yang, Shiqi Wang, Yuming Fang, Yue Wang, and Jiaying Liu.
\newblock From fidelity to perceptual quality: A semi-supervised approach for low-light image enhancement.
\newblock In {\em IEEE/CVF Conference on Computer Vision and Pattern Recognition (CVPR)}, June 2020.

\bibitem[\protect\citeauthoryear{Yang \bgroup \em et al.\egroup }{2021}]{LOLv2}
Wenhan Yang, Wenjing Wang, Haofeng Huang, Shiqi Wang, and Jiaying Liu.
\newblock Sparse gradient regularized deep retinex network for robust low-light image enhancement.
\newblock volume~30, pages 2072--2086, 2021.

\bibitem[\protect\citeauthoryear{Yi \bgroup \em et al.\egroup }{2021a}]{yi2021structure}
Qiaosi Yi, Juncheng Li, Qinyan Dai, Faming Fang, Guixu Zhang, and Tieyong Zeng.
\newblock Structure-preserving deraining with residue channel prior guidance.
\newblock In {\em ICCV}, pages 4238--4247, 2021.

\bibitem[\protect\citeauthoryear{Yi \bgroup \em et al.\egroup }{2021b}]{yi2021efficient}
Qiaosi Yi, Juncheng Li, Faming Fang, Aiwen Jiang, and Guixu Zhang.
\newblock Efficient and accurate multi-scale topological network for single image dehazing.
\newblock {\em IEEE Transactions on Multimedia}, 24:3114--3128, 2021.

\bibitem[\protect\citeauthoryear{Zamir \bgroup \em et al.\egroup }{2021}]{zamir2021multi}
Syed~Waqas Zamir, Aditya Arora, Salman Khan, Munawar Hayat, Fahad~Shahbaz Khan, Ming-Hsuan Yang, and Ling Shao.
\newblock Multi-stage progressive image restoration.
\newblock In {\em Proceedings of the IEEE/CVF conference on computer vision and pattern recognition}, pages 14821--14831, 2021.

\bibitem[\protect\citeauthoryear{Zamir \bgroup \em et al.\egroup }{2022a}]{zamir2022restormer}
Syed~Waqas Zamir, Aditya Arora, Salman Khan, Munawar Hayat, Fahad~Shahbaz Khan, and Ming-Hsuan Yang.
\newblock Restormer: Efficient transformer for high-resolution image restoration.
\newblock In {\em Proceedings of the IEEE/CVF conference on computer vision and pattern recognition}, pages 5728--5739, 2022.

\bibitem[\protect\citeauthoryear{Zamir \bgroup \em et al.\egroup }{2022b}]{Restormer}
Syed~Waqas Zamir, Aditya Arora, Salman Khan, Munawar Hayat, Fahad~Shahbaz Khan, and Ming-Hsuan Yang.
\newblock Restormer: Efficient transformer for high-resolution image restoration.
\newblock In {\em CVPR}, 2022.

\bibitem[\protect\citeauthoryear{Zeng \bgroup \em et al.\egroup }{2020}]{3DLUT}
Hui Zeng, Jianrui Cai, Lida Li, Zisheng Cao, and Lei Zhang.
\newblock Learning image-adaptive 3d lookup tables for high performance photo enhancement in real-time.
\newblock {\em IEEE Transactions on Pattern Analysis and Machine Intelligence}, 44(04):2058--2073, 2020.

\bibitem[\protect\citeauthoryear{Zeyde \bgroup \em et al.\egroup }{2012}]{zeyde2012single}
Roman Zeyde, Michael Elad, and Matan Protter.
\newblock On single image scale-up using sparse-representations.
\newblock In {\em Curves and Surfaces: 7th International Conference, Avignon, France, June 24-30, 2010, Revised Selected Papers 7}, pages 711--730. Springer, 2012.

\bibitem[\protect\citeauthoryear{Zhang \bgroup \em et al.\egroup }{2018a}]{zhang2018rcan}
Yulun Zhang, Kunpeng Li, Kai Li, Lichen Wang, Bineng Zhong, and Yun Fu.
\newblock Image super-resolution using very deep residual channel attention networks.
\newblock In {\em ECCV}, 2018.

\bibitem[\protect\citeauthoryear{Zhang \bgroup \em et al.\egroup }{2018b}]{zhang2018residual}
Yulun Zhang, Yapeng Tian, Yu~Kong, Bineng Zhong, and Yun Fu.
\newblock Residual dense network for image super-resolution.
\newblock In {\em CVPR}, 2018.

\bibitem[\protect\citeauthoryear{Zhang \bgroup \em et al.\egroup }{2019}]{KinD}
Yonghua Zhang, Jiawan Zhang, and Xiaojie Guo.
\newblock Kindling the darkness: A practical low-light image enhancer.
\newblock In {\em Proceedings of the 27th ACM International Conference on Multimedia}, MM '19, pages 1632--1640, New York, NY, USA, 2019. ACM.

\bibitem[\protect\citeauthoryear{Zhen \bgroup \em et al.\egroup }{2024}]{a:35}
Zou Zhen, Yu~Hu, and Zhao Feng.
\newblock Freqmamba: Viewing mamba from a frequency perspective for image deraining.
\newblock {\em arXiv preprint arXiv:2404.09476}, 2024.

\bibitem[\protect\citeauthoryear{Zheng \bgroup \em et al.\egroup }{2022}]{SICE-Mix}
Shen Zheng, Yiling Ma, Jinqian Pan, Changjie Lu, and Gaurav Gupta.
\newblock Low-light image and video enhancement: A comprehensive survey and beyond.
\newblock {\em arXiv preprint arXiv:2212.10772}, 2022.

\bibitem[\protect\citeauthoryear{Zhou \bgroup \em et al.\egroup }{2022}]{LEDNet}
Shangchen Zhou, Chongyi Li, and Chen~Change Loy.
\newblock Lednet: Joint low-light enhancement and deblurring in the dark.
\newblock In {\em ECCV}, 2022.

\bibitem[\protect\citeauthoryear{Zhou \bgroup \em et al.\egroup }{2023}]{zhou2023srformer}
Yupeng Zhou, Zhen Li, Chun-Le Guo, Song Bai, Ming-Ming Cheng, and Qibin Hou.
\newblock Srformer: Permuted self-attention for single image super-resolution.
\newblock {\em arXiv preprint arXiv:2303.09735}, 2023.

\bibitem[\protect\citeauthoryear{Zhu \bgroup \em et al.\egroup }{2024}]{a:30}
Lianghui Zhu, Bencheng Liao, Qian Zhang, Xinlong Wang, Wenyu Liu, and Xinggang Wang.
\newblock Vision mamba: Efficient visual representation learning with bidirectional state space model.
\newblock {\em arXiv preprint arXiv:2401.09417}, 2024.

\end{thebibliography}

\end{document}